\documentclass[10pt,twocolumn,letterpaper]{article}

\usepackage[pagenumbers]{cvpr} % To force page numbers, e.g. for an arXiv version

\definecolor{cvprblue}{rgb}{0.21,0.49,0.74}
\usepackage[pagebackref,breaklinks,colorlinks,allcolors=cvprblue]{hyperref}

\usepackage{tabularx}

\def\confName{CVPR}
\def\confYear{2027}

\title{Rank2Replace: VLM-Ranked Slot Replacement\\
for Out-of-Distribution Generalization}

\author{
 \textbf{Bo-Han Feng\textsuperscript{1}},
 \textbf{Keng-Te Liao\textsuperscript{2}},
 \textbf{Shou-De Lin\textsuperscript{1,3}}
\\
 National Taiwan University\textsuperscript{1}, 
 National Tsing Hua University\textsuperscript{2}, \\
 National Taiwan University AI Center of Research Excellence\textsuperscript{3}
\\
 \small
 \texttt{r14922016@ntu.edu.tw, ktliao@stat.nthu.edu.tw, sdlin@csie.ntu.edu.tw} \\
}

\begin{document}
\maketitle
\begin{abstract}
Spurious correlations can give image predictors high in-distribution accuracy but poor generalization under distribution shift. We introduce \textbf{Rank2Replace}, which transfers knowledge of task-relevant evidence from pretrained vision--language models (VLMs) to small predictors through training-time augmentation for out-of-distribution (OOD) generalization. The VLM ranks object-centric slots by task relevance and guides rank-dependent replacement with same-rank slot embeddings from another image. This augmentation retains the original image's label, encouraging reliance on preserved task evidence as incidental cues change, and supports combination with existing OOD methods. Across four datasets and six splits under shared frozen slot embeddings, the default cross-entropy configuration reaches macro-averaged average-group and worst-group accuracies of $80.43\%$ and $65.28\%$ without environment labels or dataset-specific replacement tuning. Slot corruption improves both macro-averaged metrics for all fourteen OOD training methods. Combining it with slot ranking and IRM reaches $81.96\%$ and $72.68\%$. Controlled ablations support the ranking and replacement design, with gains across five ranking VLMs and six extractor--predictor combinations. Permutation-invariant predictors further improve both macro-averaged metrics over their unaugmented counterparts while using VLM guidance only during training for efficient inference.
\end{abstract}

\section{Introduction}
\label{sec:introduction}

Image recognition models often exploit correlations between class labels and incidental cues, such as backgrounds and textures, induced by dataset biases~\cite{5995347,xiao2021noise,geirhos2018imagenettrained}. When these correlations change, high in-distribution accuracy can conceal severe failures on underrepresented groups. Consequently, robust out-of-distribution (OOD) generalization requires both preserving task evidence and reducing reliance on incidental cues.

Conventional robustness methods primarily address the latter goal. Environment- and group-aware methods, such as Invariant Risk Minimization (IRM)~\cite{arjovsky2020invariantriskminimization}, use observed or inferred partitions to reduce reliance on incidental cues~\cite{pmlr-v139-creager21a,labonte2023towards}, while other methods intervene through errors, gradients, or interpolation~\cite{pmlr-v139-liu21f,10.1007/978-3-030-58536-5_8,zhang2018mixup,zhou2021domain}. These mechanisms act indirectly, without explicitly identifying the task-relevant regions whose evidence should be preserved.

Vision--language models (VLMs) offer a way to make task evidence explicit. By connecting images with language~\cite{pmlr-v139-radford21a}, they can identify task-relevant regions to guide attention, representations, or augmentation~\cite{Petryk_2022_CVPR,pmlr-v328-liu26a,wong2026slidemix}. Several methods already use semantic priors, such as VLMs, to improve OOD robustness. For example, B2T derives group descriptions from predictor errors~\cite{Kim_2024_CVPR}, while RaVL uses discovered spurious regions for VLM fine-tuning~\cite{NEURIPS2024_95c6ae3f}. ASPIRE instead identifies spurious content and synthesizes training images without it~\cite{ghosh-etal-2024-aspire}. These interventions depend on the biases uncovered during diagnosis, which determine the inferred supervision or content to edit. This motivates a complementary question: how can we transfer VLMs' knowledge of task-relevant evidence to small, efficient predictors for OOD generalization?

To address this question, we introduce \textbf{Rank2Replace}, a new paradigm for VLM-guided OOD augmentation. To the best of our knowledge, it is the first method to translate instance-specific VLM task-relevance rankings into rank-dependent, cross-image replacement of object-centric slot embeddings for OOD image classification. A frozen extractor~\cite{didolkar2025on} decomposes the full image into slot images and corresponding slot embeddings that may capture objects, parts, or background. A VLM then ranks the slot images jointly with embeddings by fixed task-evidence criteria. This process preserves high-ranking slots containing task evidence. Low-ranking slots, potentially carrying incidental cues, can be replaced with same-ranked slots from other images, thereby disrupting spurious correlations between incidental cues and labels. Rule-based or VLM-derived schedules decide replacement probability for each rank. Finally, the small predictor trained on ranked slot embeddings is encouraged to rely on preserved task evidence as incidental cues change.

Across four datasets and six splits~\cite{Sagawa*2020Distributionally,liu2015faceattributes,liang2022metashift,10205329}, Rank2Replace leads diverse OOD methods with $65.28\%$ macro-averaged worst-group accuracy under shared frozen slot embeddings. Combined with IRM, Rank2Replace can even reach a higher worst-group accuracy of $72.68\%$.

\begin{samepage}
Our contributions are:
\begin{itemize}
  \item \textbf{A new paradigm leveraging VLM priors for OOD generalization.} By transferring VLM knowledge through our slot-level ranking method, Rank2Replace significantly outperforms prior works without requiring environment or group labels.
  \item \textbf{Seamless compatibility with existing methods.} When environment or group labels are available, our framework can easily integrate with existing OOD approaches to yield further performance gains.
  \item \textbf{Generality and practicality.} Rank2Replace remains effective across five VLM families and six extractor–predictor combinations, with permutation-invariant adaptation adding minimal inference latency.
\end{itemize}
\end{samepage}

\section{Related Work}
\label{sec:related_work}

\paragraph{Robust objectives and generic augmentation.}
Robust learning includes invariant prediction, worst-group risk minimization, domain alignment, class--environment regularization, and balanced refitting~\cite{arjovsky2020invariantriskminimization,Sagawa*2020Distributionally,10.1007/978-3-319-49409-8_35,ICLR2026_fd512d61,kirichenko2023last}. Reweighting, feature suppression, representation regularization, and flatness-based training offer complementary strategies~\cite{pmlr-v139-liu21f,10.1007/978-3-030-58536-5_8,pmlr-v258-wen25a,Wang_2023_CVPR,NEURIPS2021_bcb41ccd}. Mixup and MixStyle augment inputs or feature statistics~\cite{zhang2018mixup,zhou2021domain}.

\paragraph{Cross-image feature composition.}
SwapMix~\cite{gupta2022swapmix} uses ground-truth question reasoning to identify irrelevant object features in VQA. Class swaps favor semantically similar categories, while attribute swaps retain the object category. Rank2Replace matches pretrained slots by task-relevance rank, without question programs or category/attribute matching. An extractor partitions the full image into slots, covering objects, parts, and background. A  VLM then supplies an image-specific slot ranking from a fixed task description, without human per-image relevance annotations. This permits replacement beyond a predefined object inventory, including slots containing both task evidence and context. Slot-Mixup aggregates whole-slide features into slots and interpolates slots and labels~\cite{keum2023slotmixup}. SlideMix uses a VLM to select diagnostically relevant tissue features and supplies VLM soft labels for shuffled mixtures~\cite{wong2026slidemix}. Rank2Replace replaces whole slot embeddings while retaining labels, avoiding the need to assign mixed training targets. The full task-relevance ranking specifies the default predictor's slot positions and replacement-schedule indices. This couples semantic alignment to slot replacement, including graded schedules that retain lower-ranked slots with nonzero probability. Lee et al.~\cite{NEURIPS2021_d360a502} learn intrinsic and bias encoders, then randomly permute bias vectors across a minibatch while retaining recipient intrinsic factors and labels. BiaSwap~\cite{9710845} transfers bias attributes through CAM-guided image translation with a swapping autoencoder. Rank2Replace obtains its components from a pretrained extractor without learning a binary intrinsic/bias factorization or training a swapping generator.

\paragraph{Semantic guidance for robustness.}
GALS supervises attention with language-grounded task-relevance maps~\cite{Petryk_2022_CVPR}. SupER uses superclass-guided attention alignment and representation disentanglement~\cite{pmlr-v328-liu26a}. These are precedents for positive task guidance. Rank2Replace applies this guidance through ordered feature composition without requiring attention alignment. Augmented embeddings can feed different predictor architectures and training objectives. RaVL discovers spurious region clusters and derives subgroup information for region-aware VLM fine-tuning, including encouragement of relevant regions~\cite{NEURIPS2024_95c6ae3f}. Its intervention therefore depends on corpus-level bias discovery and subgroup-derived training signals. Our default pipeline ranks each image from a task description, without inferred group partitions or VLM fine-tuning. B2T extracts keywords from captions of misclassified images to derive group descriptions~\cite{Kim_2024_CVPR}. Yang et al. detect candidate spurious attributes and retain practitioner validation before debiasing~\cite{pmlr-v202-yang23j}. These approaches diagnose biases from model behavior across examples. Rank2Replace orders each image's visible task evidence without using downstream prediction errors. LBC and SPUME use discovered class--attribute correlations for relabeling or meta-learning~\cite{10.24963/ijcai.2024/619,10.1145/3637528.3672006}. DISC and ASPIRE target detected spurious concepts through augmentation or synthesis~\cite{3618408.3619982,ghosh-etal-2024-aspire}. LADS, CLIP the Gap, and P{\O}DA use domain descriptions~\cite{dunlap2023using,Vidit_2023_CVPR,Fahes_2023_ICCV}. FAMix mines local style banks from random prompts~\cite{Fahes_2024_FAMix}. Rank2Replace exchanges slot embeddings across images by task relevance without domain descriptions, style banks, or prior bias discovery, and is compatible with different training pipelines.

\paragraph{Representations and predictive interfaces.}
Object-centric methods learn concept dictionaries, condition or fuse slots with language, and build object--concept--relation representations~\cite{pmlr-v235-arefin24a,Didolkar_2025_CTRLO,Li_2024_DOC,Tang_2025_OCRT}. Rank2Replace applies task-relevance rankings to already extracted slots. Concept bottleneck methods predict through explicit concepts~\cite{pmlr-v119-koh20a,oikarinen2023labelfree,yang2023language}. Richer concept embeddings and residual pathways mitigate the information loss of incomplete scalar concept vocabularies~\cite{NEURIPS2022_867c0682,yuksekgonul2023posthoc,10656628,NEURIPS2025_db251964}. This evolution justifies Rank2Replace's prediction from native slot embeddings.

\begin{figure*}[t]
\includegraphics[width=1.0\linewidth]{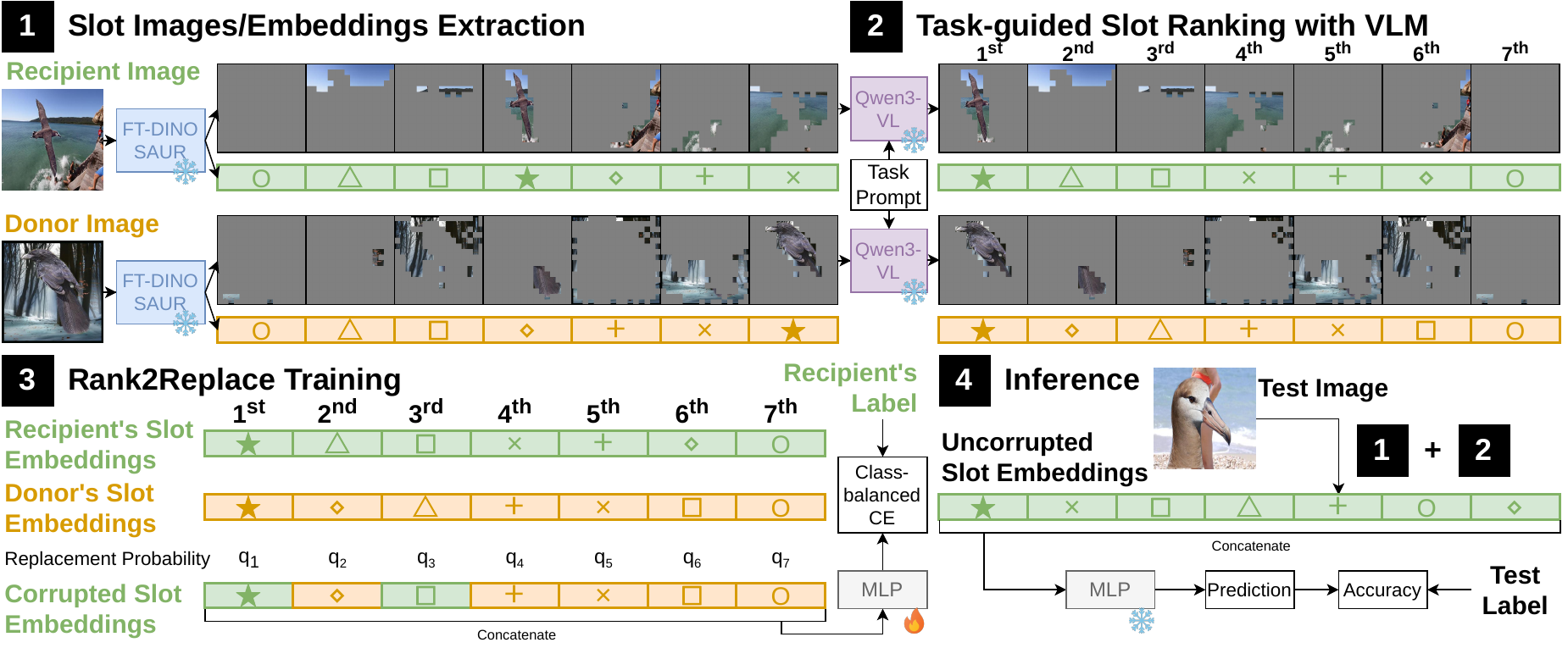}
  \caption{
  Rank2Replace training and inference with rank-dependent corruption, illustrated with the Waterbirds dataset.
  }
  \label{fig:method}
\end{figure*}

\section{Method}
\label{sec:method}

Rank2Replace uses VLM priors to rank slots before cross-image replacement for data augmentation (Fig.~\ref{fig:method}). Rankings define the positions of all slots and their donor correspondence. Rank-dependent probabilities determine which recipient embeddings are replaced.

\subsection{Object-centric Representations}
\label{sec:slot_representations}

Given a training set $\mathcal{D}=\{(x_i,y_i)\}_{i=1}^{N}$ with $y_i\in\{1,\ldots,C\}$, a frozen, pretrained object-centric extractor produces $K\geq2$ slot embeddings $h_{i,s}\in\mathbb{R}^{d}$ and corresponding region masks. Our hard masks assign every image patch to a slot, partitioning the full input image (App.~\ref{app:implementation}). The original slot index $s$ has no prescribed semantic order. Each mask retains its corresponding region and fills the remaining pixels with a fixed gray value, producing a slot image $v_{i,s}$. Slot images provide visible evidence for ranking, and slot embeddings provide the predictive representation.

\subsection{Task-guided Slot Ranking}
\label{sec:slot_ranking}

A frozen VLM ranks each image's nonempty slot images using a task description and task-evidence criteria. A shared prompt prioritizes pure task evidence, then evidence mixed with unrelated content, then regions without task evidence. The prompt, task descriptions, and evidence criteria are human-authored before predictor training (App.~\ref{app:vlm_prompts}). VLM ranking only uses slot images without ground-truth labels. Empty slots follow in a fixed random order. Evidence annotations support ranking effectiveness. (App.~\ref{app:ranking_evidence}).

Let $\pi_i(r)$ denote the original slot index at rank $r$, with $r=1$ being the most relevant. The ranked representation is
\begin{equation}
\begin{aligned}
z_{i,r} &= T_r\!\left(h_{i,\pi_i(r)}\right),\\
z_i &= \operatorname{concat}(z_{i,1},\ldots,z_{i,K})
      \in\mathbb{R}^{Kd},
\end{aligned}
\label{eq:ranked_slots}
\end{equation}
where $T_r:\mathbb{R}^{d}\to\mathbb{R}^{d}$ is fixed preprocessing shared across images at rank $r$: the identity or standardization estimated only from training data. The same permutation orders the slot images and embeddings. For the default predictor, the full permutation specifies input coordinates and same-rank donor correspondence. It also indexes replacement probabilities. Equal ranks need not share an object category.

\subsection{Rank-dependent Slot Corruption}
\label{sec:slot_corruption}

A schedule assigns $(q_1,\ldots,q_K)\in[0,1]^K$, either by a rule on rank or from rank-wise training-set aggregates of VLM relevance scores (App.~\ref{app:prompt_policy}). For a minibatch of $B$ images, a uniformly random permutation $\sigma$ assigns donor $\sigma(i)$ to recipient $i$. Pairing is independent of class labels and allows self-pairs. We independently sample
\begin{equation}
m_{i,r}\sim\operatorname{Bernoulli}(q_r),\qquad r=1,\ldots,K,
\label{eq:replacement_schedule}
\end{equation}
and form corrupted slot embeddings by replacement
\begin{equation}
\begin{aligned}
\widetilde z_{i,r}
  &= (1-m_{i,r})z_{i,r}+m_{i,r}z_{\sigma(i),r},\\
\widetilde z_i
  &= \operatorname{concat}(\widetilde z_{i,1},\ldots,\widetilde z_{i,K}).
\end{aligned}
\label{eq:slot_corruption}
\end{equation}
Replacement uses whole, same-rank embeddings from one shared donor per recipient as default, retaining a common source for inserted content. Donor embeddings remain native extractor outputs after the same preprocessing. We resample $\sigma$ for each training batch.

Our default \emph{Top-preserving} schedule fixes $q_1=0$ and $q_r=1$ for $r>1$ across datasets. Section~\ref{sec:replacement_schedules} compares rule-based and VLM-derived schedules.

% Queue the first experimental table one page before its discussion.
\begin{table*}[t]
\centering
\caption{\textbf{Global image embeddings and end-to-end training.} All conditions start from the same FT-DINOSAUR encoder checkpoint. Bold marks the default design and highest macro-averaged accuracies. Settings: App.~\ref{app:global_image_embeddings}.}
\label{tab:global_macro}
\small
\setlength{\tabcolsep}{3pt}
\renewcommand{\arraystretch}{1.06}
\begin{tabular*}{\textwidth}{@{\extracolsep{\fill}}lllrr@{}}
\toprule
Representation / input & Encoder & Predictor / training & Average-group & Worst-group \\
\midrule
Global image embedding & Frozen & Linear / CE & $75.60 \pm 0.09$ & $54.13 \pm 0.33$ \\
 & Frozen & MLP / CE & $75.56 \pm 0.56$ & $53.89 \pm 1.82$ \\
 & Updated & Linear / CE & $79.41 \pm 1.01$ & $59.75 \pm 3.06$ \\
 & Updated, then frozen & Linear / AFR & $78.93 \pm 0.93$ & $58.87 \pm 2.73$ \\
 & Updated & Linear / SagNet & $80.46 \pm 1.06$ & $62.07 \pm 4.26$ \\
\midrule
Original slot order & Frozen & Linear / CE & $72.14 \pm 0.32$ & $49.43 \pm 1.18$ \\
Original slot order & Frozen & MLP / CE & $71.72 \pm 0.66$ & $47.77 \pm 3.02$ \\
Slot ranking & Frozen & MLP / CE & $75.96 \pm 0.29$ & $55.48 \pm 0.63$ \\
\textbf{Rank2Replace} & \textbf{Frozen} & \textbf{MLP / CE} & $80.43 \pm 0.33$ & $65.28 \pm 0.81$ \\
Rank2Replace & Updated & MLP / CE & $\mathbf{81.72} \pm 0.54$ & $\mathbf{67.52} \pm 2.24$ \\
\bottomrule
\end{tabular*}
\end{table*}

\subsection{Training and Inference}
\label{sec:training_inference}

We retain recipient class labels to form $\widetilde{\mathcal{B}}=\{(\widetilde z_i,y_i)\}_{i=1}^{B}$ and optimize a predictor $f_\theta$ with training objective $\mathcal{L}_{\mathrm{task}}$,
\begin{equation}
\min_{\theta}\;
\mathbb{E}_{\mathcal{B},\sigma,m}
\left[\mathcal{L}_{\mathrm{task}}
      (\theta;\widetilde{\mathcal{B}})\right].
\label{eq:training_objective}
\end{equation}
The default design uses class-balanced cross-entropy (CE), and trains only the downstream predictor. Slot corruption supports other objectives with their own supervision requirements or predictor architectures.

With the default MLP predictor, validation and inference use uncorrupted slot embeddings with the same ranking and preprocessing. Permutation-invariant predictors instead use shared preprocessing across slots and require VLM ranking only for training-time replacement (Sec.~\ref{sec:vlm_free_inference}).

\section{Experiments}
\label{sec:experiments}

\subsection{Setup}
\label{sec:setup}

\paragraph{Datasets and evaluation.}
We evaluate Waterbirds, CelebA, MetaShift, and three NICO++ splits (App.~\ref{app:datasets}). We report average-group and worst-group accuracy over class $\times$ environment groups for the first three datasets and target-environment groups for NICO++. Macro averages weight the four datasets equally after averaging the three NICO++ splits. We report mean $\pm$ sample SD over 3 seeds. Accuracies and their absolute differences use \%.

\paragraph{Shared protocol and baselines.}
To isolate downstream robustness mechanisms, methods share frozen FT-DINOSAUR slot embeddings and an MLP unless varied in an ablation, with $K=7$ by default (App.~\ref{app:implementation}). Default ranking uses Qwen3-VL-8B-Instruct~\cite{bai2025qwen3vltechnicalreport}. In the overall comparison, Rank2Replace, Grouped LAAT, and hard removal use ranked slot embeddings. Other methods use the original slot order. Grouped LAAT aligns slot importance with the ranking~\cite{10.1007/978-3-032-08327-2_4}, and hard removal retains a validation-selected number of top-ranked slots (App.~\ref{app:baselines}). IRM, Group DRO, CORAL, SCER, and DFR use Qwen3-VL environment labels by default.

\paragraph{Model selection.}
Predictor training uses the same method-specific search spaces. Models and hyperparameters are selected by validation balanced accuracy plus macro F1 (App.~\ref{app:evaluation}). Default Rank2Replace fixes Top-preserving across datasets without a robustness penalty to tune.

\subsection{Main Results}
\label{sec:main_results}

We first examine the necessity of slot embeddings, then compare robustness and compatibility under shared frozen slot embeddings. Detailed settings are in App.~\ref{app:detailed_results}.

\begin{figure*}[t]
  \centering
  \includegraphics[width=\textwidth]{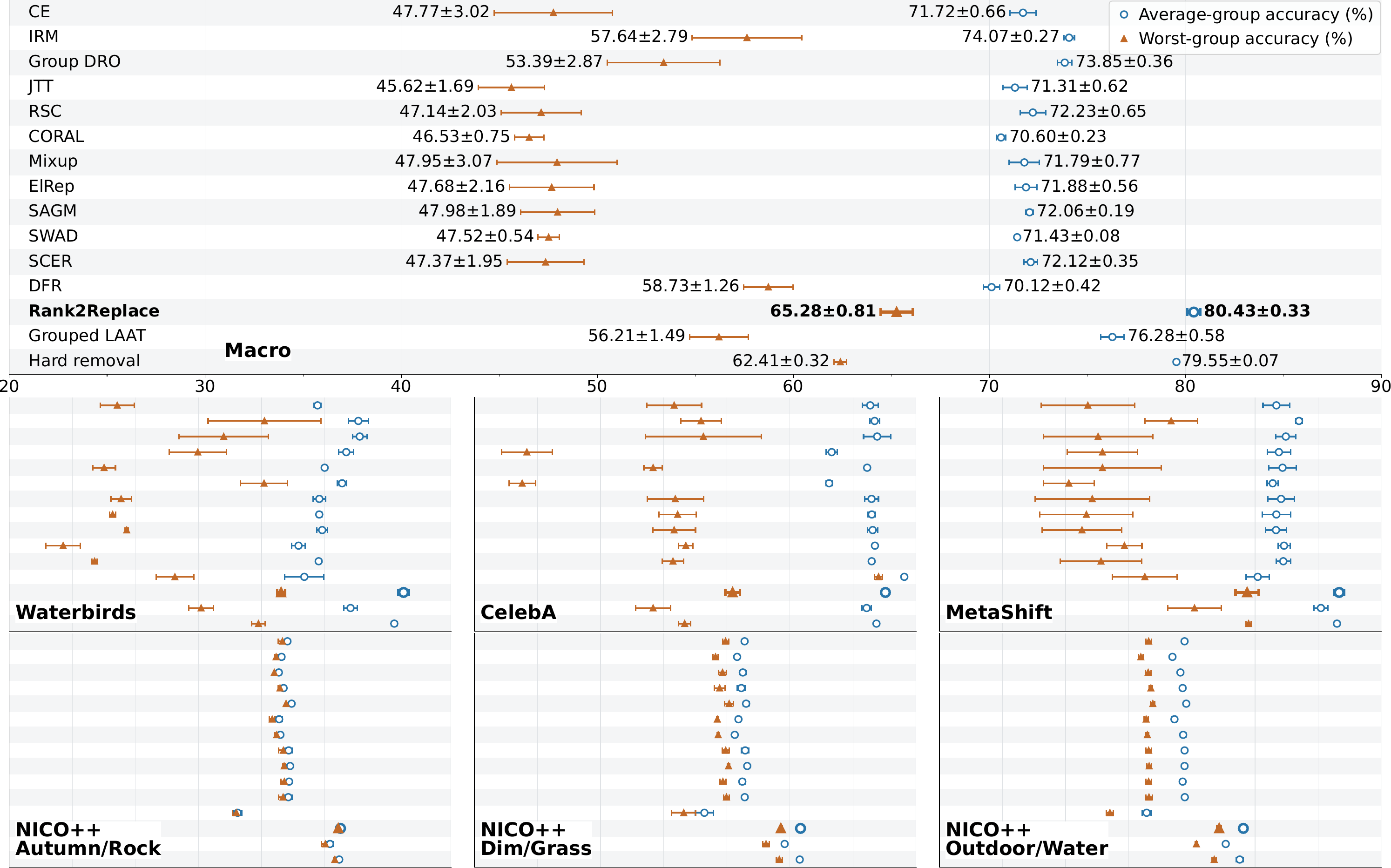}
  \caption{\textbf{Overall performance.} Rank2Replace leads both macro-averaged metrics (top). Individual splits (below) show exceptions for DFR on CelebA and hard removal in MetaShift worst-group accuracy. Split-level numerical results are in Tables~\ref{tab:overall_context} and~\ref{tab:overall_nico}.}
  \label{fig:baseline_performance}
\end{figure*}

\subsubsection{Global Embeddings and End-to-End Training}
\label{sec:global_image_embeddings}

Table~\ref{tab:global_macro} compares mean-pooled global image embeddings and slot embeddings from the same initial encoder. With a frozen encoder and an MLP, Rank2Replace reaches $80.43\%$ average-group and $65.28\%$ worst-group accuracy, exceeding global image embeddings by $4.87\%$ and $11.39\%$. Ranking alone improves average-/worst-group accuracy over original-order CE by $4.24\%$/$7.71\%$, before replacement adds $4.47\%$/$9.80\%$. These controls support slot ranking and replacement for prediction and augmentation.

The advantage also holds against methods that update the global-image encoder. Frozen Rank2Replace exceeds AFR~\cite{pmlr-v202-qiu23c} in both metrics and SagNet~\cite{Nam2021ReducingDG} in worst-group accuracy ($65.28\%$ versus $62.07\%$), with nearly identical average-group accuracy ($80.43\%$ versus $80.46\%$).

Jointly updating the encoder, Slot Attention, and MLP with fixed Qwen3-VL rankings raises Rank2Replace to $81.72\%$/$67.52\%$, improving over the frozen default by $1.29\%$/$2.24\%$ and exceeding both AFR and SagNet in both metrics. These are separate pipeline comparisons. Subsequent comparisons use shared frozen slot embeddings.

\subsubsection{Overall Performance}
\label{sec:overall_performance}
Rank2Replace exceeds all fourteen baselines in both macro-averaged metrics, reaching $80.43\%$ average-group and $65.28\%$ worst-group accuracy (Fig.~\ref{fig:baseline_performance}). These exceed CE by $8.71\%$ and $17.51\%$. It leads average-group accuracy on five of six splits and worst-group accuracy on four. The exceptions are DFR on CelebA ($88.13\%$/$84.05\%$ versus $85.13\%$/$60.93\%$) and hard removal on MetaShift worst-group accuracy ($68.98\%$ versus $68.75\%$).

On Waterbirds and MetaShift, worst-group gains over CE are $25.96\%$ and $25.23\%$. Rank2Replace exceeds hard removal by $0.88\%$/$2.87\%$ in the two macro-averaged metrics, despite hard removal's split-specific cutoff tuning. It also leads both metrics for $K=5$--$9$ (App.~\ref{app:slot_count}). Across a $3\times3$ learning-rate and weight-decay sweep, default Rank2Replace even exceeds every IRM, Group DRO, and DFR configuration in both metrics (App.~\ref{app:optimizer_sensitivity}).

\begin{figure*}[t]
  \centering
  \includegraphics[width=0.96\textwidth]{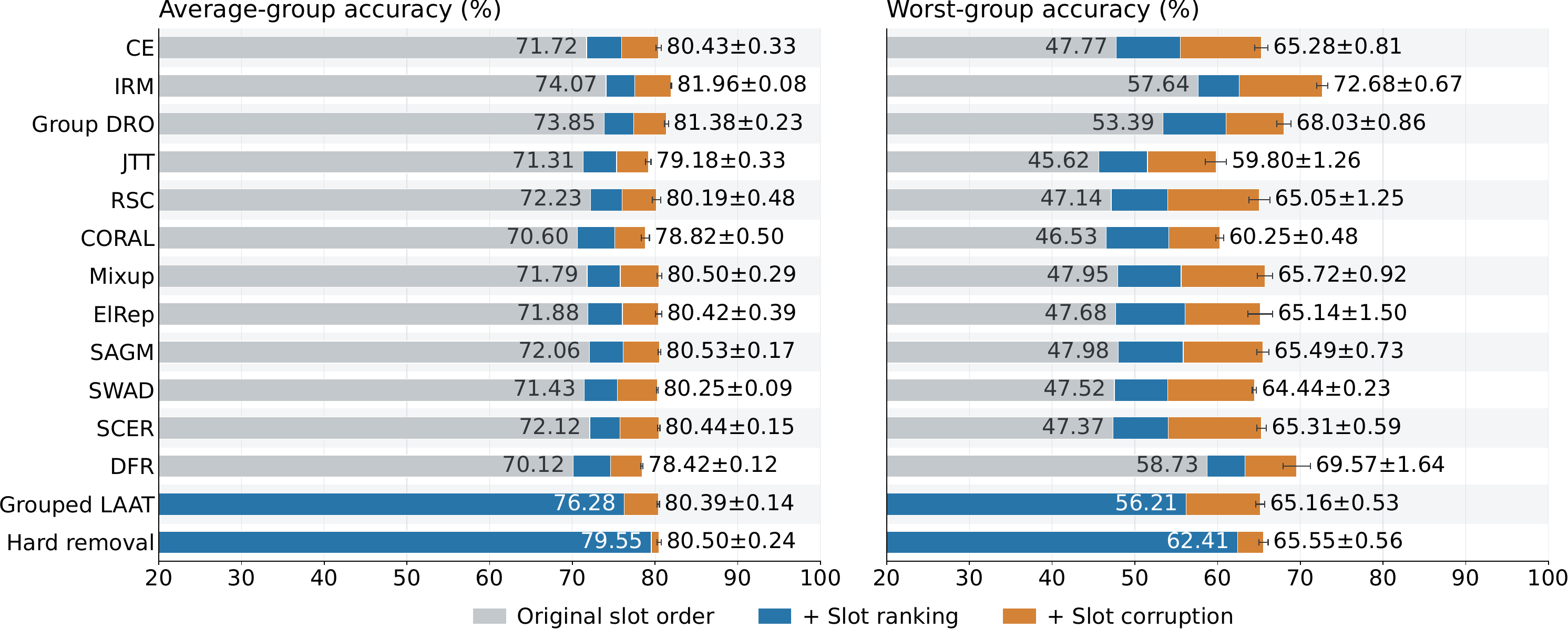}
  \caption{\textbf{Slot ranking and slot corruption across training methods.} Gray bars use the original slot order, blue adds slot ranking, and orange adds slot corruption. Grouped LAAT and hard removal already use slot ranking. CE with both interventions is default Rank2Replace. Error bars refer to models trained with slot corruption. Numerical results: Tables~\ref{tab:compatibility_average} and~\ref{tab:compatibility_worst}.}
  \label{fig:compatibility_macro}
\end{figure*}

\begin{figure*}[t]
  \centering
  \includegraphics[width=0.96\textwidth]{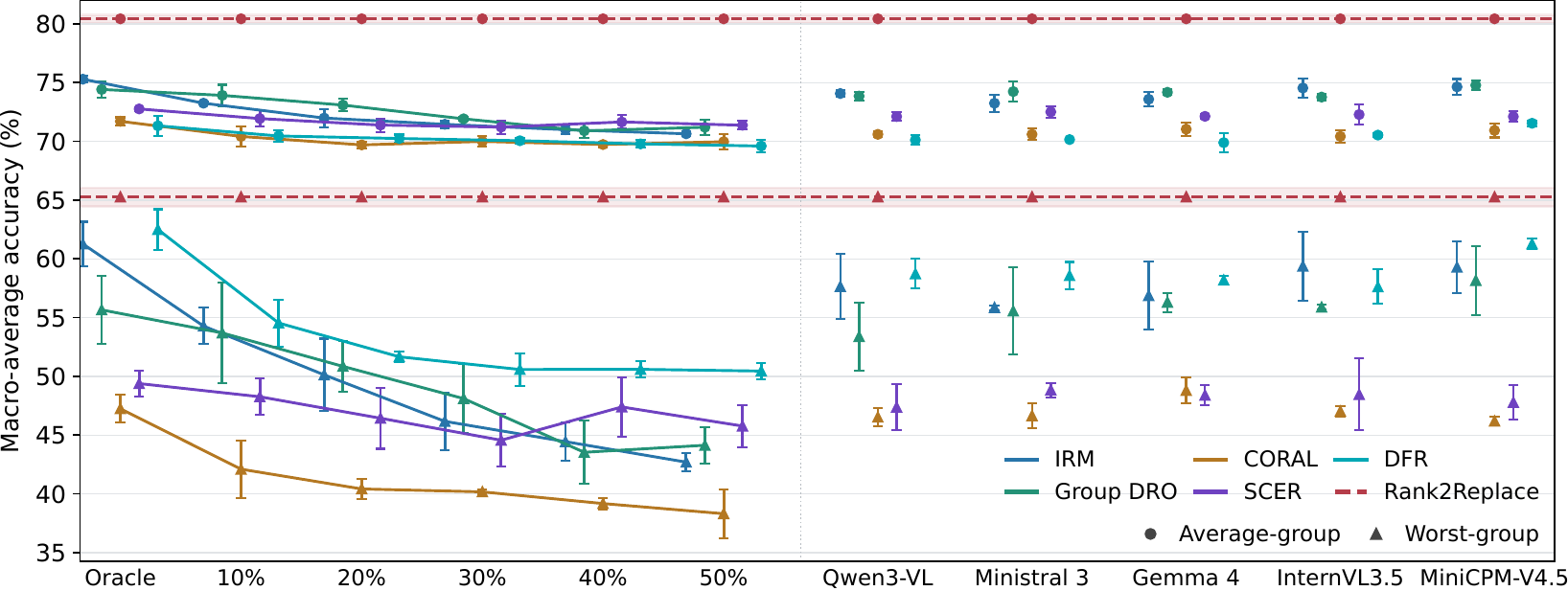}
  \caption{\textbf{Effect of environment labels.} Baselines use oracle labels, 10\%--50\% synthetic noise, or labels from five VLMs. Circles and triangles show average- and worst-group accuracy. Red lines and bands show the fixed Rank2Replace reference. Results: Table~\ref{tab:environment_quality}.}
  \label{fig:environment_downstream}
\end{figure*}

\subsubsection{Contributions and Compatibility}
\label{sec:ranking_compatibility}
Slot ranking improves both macro-averaged metrics for all twelve methods with an original-order control, and slot corruption further improves all fourteen methods in Fig.~\ref{fig:compatibility_macro}. Across the twelve complete comparisons, ranking increases worst-group accuracy by $4.56\%$--$8.38\%$, and corruption adds $6.15\%$--$11.26\%$.

Rank2Replace complements environment-aware objectives. Combined with IRM, it reaches $81.96\%$ average-group and $72.68\%$ worst-group accuracy, the highest means among the tested combinations and improvements of $1.53\%$ and $7.40\%$ over default Rank2Replace. Group DRO reaches $81.38\%$/$68.03\%$, also improving both metrics. DFR reaches $78.42\%$/$69.57\%$, exchanging lower average-group accuracy for higher worst-group accuracy. These combinations retain their environment-label requirements.

Slot corruption raises worst-group accuracy by $8.95\%$ for Grouped LAAT and $3.14\%$ for hard removal.

Rank2Replace also complements methods that already use language-derived guidance: it raises macro-averaged worst-group accuracy beyond ranking alone by $7.89\%$ for DRO-B2T and $16.67\%$ for SPUME (App.~\ref{app:semantic_compatibility}). These additional pipelines use CLIP-derived group supervision or attribute-guided meta-learning. Conversely, replacing Qwen3-VL ranking with random or mask-area orderings substantially reduces performance under Top-preserving, indicating that slot ranking is essential for slot corruption (App.~\ref{app:ranking_controls}).

\begin{figure*}[t]
\centering
\includegraphics[width=0.9\textwidth]{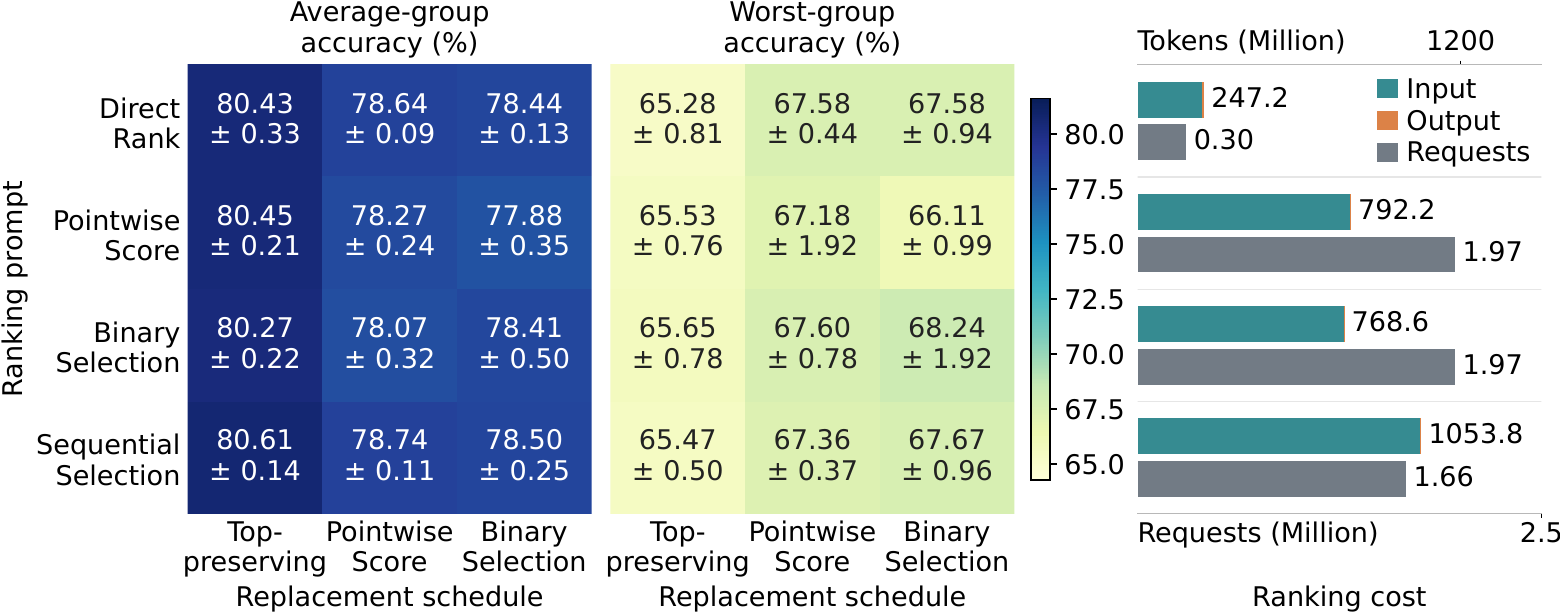}
\caption{\textbf{Ranking prompts and replacement schedules.} Heatmaps cross four ranking prompts with three replacement schedules. Costs count ranking-stage tokens and requests, excluding additional training-slot scoring (App.~\ref{app:prompt_policy}).}
\label{fig:ranking_prompts_replacement_schedules}
\end{figure*}

\subsubsection{Environment-Label Quality}
\label{sec:environment_quality}
Without environment labels, Rank2Replace exceeds IRM, Group DRO, CORAL, SCER, and DFR in both macro-averaged metrics under every tested label condition (Fig.~\ref{fig:environment_downstream}). Even with oracle labels, the strongest baseline means are $75.29\%$ average-group accuracy (IRM) and $62.50\%$ worst-group accuracy (DFR), trailing by $5.14\%$ and $2.78\%$. With VLM labels, the best means are $74.78\%$ and $61.27\%$, obtained by Group DRO and DFR using MiniCPM-V4.5. Environment-inference errors alone therefore cannot explain Rank2Replace's gains.

Synthetic label noise affects worst-group accuracy more strongly than average-group accuracy for all five methods at every tested rate. At 50\% noise, declines from oracle results are $3.62\%$--$18.57\%$ and $1.38\%$--$4.64\%$, respectively. Variation across VLM label sources is smaller than each method's corresponding noise-induced declines.

\subsection{Ablation Studies}
\label{sec:ablations}

We examine VLM queries, replacement schedules, and donor construction, then architecture and inference trade-offs. Appendix~\ref{app:detailed_results} follows the same experiment order with numerical results and detailed settings. App.~\ref{app:additional_ablations} provides additional analyses.

\subsubsection{Ranking Prompts and Replacement Schedules}
\label{sec:ranking_replacement_design}
\label{sec:replacement_schedules}

Figure~\ref{fig:ranking_prompts_replacement_schedules} examines how VLM outputs guide ranking and replacement. Independent annotations show that Qwen3-VL tends to place recognizable task evidence at the front of its rankings (App.~\ref{app:ranking_evidence}). Corruption improves both macro-averaged metrics with all five tested ranking VLMs, while slot-image presentation order has little effect (App.~\ref{app:vlm_ranking}). Direct Rank uses one joint query, Pointwise Score and Binary Selection score individual slots, and Sequential Selection uses repeated queries. Direct Rank has the lowest measured ranking latency, $269$~ms/image. Top-preserving needs no auxiliary scores. Score-derived schedules reuse matching ranking outputs or require additional training-slot scoring (App.~\ref{app:prompt_policy}).

Top-preserving yields similar macro-averaged results across ranking prompts with $80.27\%$--$80.61\%$ average-group and $65.28\%$--$65.65\%$ worst-group accuracy. Sequential Selection improves on Direct Rank by $0.18\%$/$0.19\%$ while requiring $2.7\times$ the ranking latency. Within each ranking prompt, Pointwise Score and Binary Selection schedules trade $1.79\%$--$2.57\%$ average-group accuracy for $0.58\%$--$2.59\%$ worst-group gains. Their finer-grained probabilities retain lower-ranked slots with nonzero probability, at the cost of more expensive ranking or additional training-slot scoring (App.~\ref{app:controlled_schedules}). Direct Rank with Top-preserving favors low cost and higher average-group accuracy.

With fixed Direct Rank prompt, Linear and Reversed lower-rank replacement schedules reach $80.01\%$/$63.81\%$ and $79.94\%$/$64.11\%$. These comparisons vary probabilities while retaining full slot rankings and same-rank donor correspondence. With uncorrupted validation/test inputs fixed, donor replacement exceeds train-only masking by $19.55\%$/$26.68\%$ and train-only moment-matched synthetic corruption ($79.52\%$/$62.15\%$) by $0.91\%$/$3.13\%$. The latter supports native donor content beyond independent coordinatewise moment matching (App.~\ref{app:controlled_schedules}).

\begin{table}[t]
\caption{\textbf{Donor source.} Fixed Qwen3-VL ranking and Top-preserving. Bold marks the default and highest means. Setting definitions: App.~\ref{app:donor_source}.}
\label{tab:donor_source}
\centering
\small
\setlength{\tabcolsep}{3pt}
\renewcommand{\arraystretch}{1.04}
\begin{tabularx}{\columnwidth}{@{}>{\raggedright\arraybackslash}Xrr@{}}
\toprule
Setting & Average-group & Worst-group \\
\midrule
\textbf{Shared donor} & $80.43 \pm 0.33$ & $65.28 \pm 0.81$ \\
Two-donor choice & $80.57 \pm 0.42$ & $65.43 \pm 1.11$ \\
Independent donors & $80.58 \pm 0.20$ & $\mathbf{66.18} \pm 0.88$ \\
Two-donor mean & $80.47 \pm 0.29$ & $65.10 \pm 0.95$ \\
\footnotesize Two-donor mean + scale correction & $80.47 \pm 0.29$ & $65.10 \pm 0.95$ \\
\bottomrule
\end{tabularx}
\end{table}

\begin{table*}[t]
\centering
\caption{\textbf{Extractor and predictor generalizability.} Cumulative slot ranking and slot corruption across two frozen extractors and three predictors. Bold marks default choices and best means within each extractor--predictor pair. The Transformer uses positional embeddings.}
\label{tab:extractor_predictor}
\small
\setlength{\tabcolsep}{3pt}
\renewcommand{\arraystretch}{1.02}
\begin{tabular*}{\textwidth}{@{\extracolsep{\fill}}llrrrr@{}}
\toprule
 & & \multicolumn{2}{c}{\textbf{FT-DINOSAUR}} & \multicolumn{2}{c}{SPOT} \\
\cmidrule(lr){3-4}\cmidrule(lr){5-6}
Predictor & Input condition & Average-group & Worst-group & Average-group & Worst-group \\
\midrule
\textbf{MLP} & Original slot order & $71.72 \pm 0.66$ & $47.77 \pm 3.02$ & $72.90 \pm 0.32$ & $49.59 \pm 1.69$ \\
 & + Slot ranking & $75.96 \pm 0.29$ & $55.48 \pm 0.63$ & $73.75 \pm 0.33$ & $52.60 \pm 1.56$ \\
 & \textbf{+ Slot corruption} & $\mathbf{80.43} \pm 0.33$ & $\mathbf{65.28} \pm 0.81$ & $\mathbf{77.73} \pm 0.27$ & $\mathbf{62.17} \pm 0.46$ \\
\midrule
Linear & Original slot order & $72.14 \pm 0.32$ & $49.43 \pm 1.18$ & $73.56 \pm 0.58$ & $50.14 \pm 1.28$ \\
 & + Slot ranking & $75.27 \pm 0.36$ & $54.15 \pm 3.38$ & $74.74 \pm 0.34$ & $53.92 \pm 1.66$ \\
 & \textbf{+ Slot corruption} & $\mathbf{79.39} \pm 0.13$ & $\mathbf{62.00} \pm 1.21$ & $\mathbf{76.92} \pm 0.19$ & $\mathbf{59.76} \pm 0.81$ \\
\midrule
Transformer & Original slot order & $75.42 \pm 0.69$ & $51.96 \pm 1.86$ & $73.72 \pm 0.47$ & $51.18 \pm 1.27$ \\
 & + Slot ranking & $75.21 \pm 0.77$ & $52.14 \pm 2.42$ & $74.21 \pm 0.68$ & $53.14 \pm 1.60$ \\
 & \textbf{+ Slot corruption} & $\mathbf{79.57} \pm 0.68$ & $\mathbf{63.85} \pm 2.08$ & $\mathbf{77.01} \pm 0.14$ & $\mathbf{60.37} \pm 0.84$ \\
\bottomrule
\end{tabular*}
\end{table*}

\begin{table*}[t]
\centering
\caption{\textbf{VLM direct prediction and VLM-free inference.} Accuracy (\%) and latency (ms/image) for full-image Qwen3-VL prediction and slot-based predictors. Features include FT-DINOSAUR and, for Deep Sets, its shared slot MLP. A dash denotes an unused inference stage. Total sums stage times. Bold marks the highest accuracy and lowest total latency.}
\label{tab:vlm_free_inference}
\small
\setlength{\tabcolsep}{3.3pt}
\renewcommand{\arraystretch}{1.06}
\begin{tabular*}{\textwidth}{@{\extracolsep{\fill}}lrrrrrr@{}}
\toprule
 & \multicolumn{2}{c}{Macro-averaged accuracy (\%)} & \multicolumn{4}{c}{Latency (ms/image)} \\
\cmidrule(lr){2-3}\cmidrule(lr){4-7}
Setting & Average-group & Worst-group & Features & VLM & Predictor & Total \\
\midrule
VLM direct prediction & $80.14$ & $63.67$ & -- & $135.011$ & -- & $135.011$ \\
MLP + CE & $71.72 \pm 0.66$ & $47.77 \pm 3.02$ & $2.150$ & -- & $0.069$ & $\mathbf{2.219}$ \\
\textbf{MLP + Rank2Replace} & $80.43 \pm 0.33$ & $65.28 \pm 0.81$ & $2.150$ & $269.038$ & $0.069$ & $271.257$ \\
Deep Sets + CE & $76.52 \pm 0.59$ & $55.84 \pm 1.32$ & $3.883$ & -- & $0.057$ & $3.940$ \\
Deep Sets + Rank2Replace & $80.40 \pm 0.33$ & $\mathbf{66.21} \pm 1.03$ & $3.883$ & -- & $0.057$ & $3.940$ \\
Set Transformer + CE & $76.70 \pm 0.53$ & $54.11 \pm 1.72$ & $2.150$ & -- & $0.242$ & $2.392$ \\
Set Transformer + Rank2Replace & $\mathbf{80.65} \pm 0.20$ & $65.46 \pm 1.19$ & $2.150$ & -- & $0.242$ & $2.392$ \\
\bottomrule
\end{tabular*}
\end{table*}

\subsubsection{Donor Source}
\label{sec:donor_source}

Having established the value of donor content, Table~\ref{tab:donor_source} varies how corrupted embeddings are constructed. One shared donor is the simplest condition, reusing a single image's lower-ranked slots for each recipient. More elaborate constructions offer modest additional gains: independent donors reach $66.18\%$ worst-group accuracy, $0.90\%$ above the default, while scale-corrected two-donor averaging reaches the highest average-group accuracy of $80.59\%$.

Scale correction improves two-donor averaging, while the shared donor remains competitive with fewer sampling and transformation steps.

\subsubsection{Extractor and Predictor Generalizability}
\label{sec:extractor_predictor}

Rank2Replace improves both macro-averaged metrics over ranking alone and the original slot order in all six extractor--predictor combinations (Table~\ref{tab:extractor_predictor}). We pair FT-DINOSAUR and SPOT~\cite{Kakogeorgiou2024SPOT} with an MLP, a linear predictor, and a Transformer, keeping ranking and replacement fixed. Beyond ranking alone, corruption improves average-group accuracy by $2.18\%$--$4.47\%$ and worst-group accuracy by $5.84\%$--$11.71\%$. Predictor details are in App.~\ref{app:extractor_predictor}.

The FT-DINOSAUR Transformer changes little with ranking alone, but corruption raises accuracy from $75.21\%$/$52.14\%$ to $79.57\%$/$63.85\%$. SPOT also improves, though its accuracies remain lower. FT-DINOSAUR's linear predictor with Rank2Replace exceeds its original-order Transformer ($79.39\%$/$62.00\%$ versus $75.42\%$/$51.96\%$).

\pagebreak[0]
\begin{samepage}
\subsubsection{VLM Direct Prediction and VLM-free Inference}
\label{sec:vlm_free_inference}

Deep Sets~\cite{3294996.3295098} and Set Transformer~\cite{lee2019set} use shared slot standardization and permutation-invariant aggregation. Rankings guide replacement only during training. Validation and test use uncorrupted, original-order slots without VLM calls (App.~\ref{app:vlm_free_inference}).

\end{samepage}

Rank2Replace improves both invariant predictors at unchanged inference cost (Table~\ref{tab:vlm_free_inference}). Set Transformer reaches $80.65\%$/$65.46\%$ at $2.392$~ms/image, and Deep Sets reaches $80.40\%$/$66.21\%$ at $3.940$~ms/image. Both exceed direct VLM prediction's macro-averaged accuracies, with $56.4\times$ and $34.3\times$ lower summed stage latency. Set Transformer has lower latency and higher average-group accuracy; Deep Sets has higher worst-group accuracy.

The default MLP requires inference-time ranking, while MLP + CE is fastest at $2.219$~ms/image but has lower accuracy. Permutation-invariant predictors retain VLM guidance during training without VLM calls at deployment.

\section{Conclusion}
\label{sec:conclusion}

Rank2Replace introduces a new paradigm for transferring VLM knowledge of task-relevant evidence to small predictors for OOD generalization. To our knowledge, it is the first method to translate instance-specific VLM task-relevance rankings into rank-dependent, cross-image replacement of object-centric slot embeddings for OOD image classification. Ranking prioritizes evidence preservation, while donor replacement discourages reliance on incidental cues without requiring environment labels or prior bias discovery. This guidance enhances both traditional and VLM-based OOD approaches, with consistent gains across ranking VLMs, extractors, and predictors. Permutation-invariant predictors further confine VLM use to training, enabling efficient inference for real-world deployment.

% The mandatory statement follows the main text and precedes references.
% It is excluded from the eight-page limit.
\clearpage
\section*{LLM Usage Statement}

We used ChatGPT with GPT-6 Pro to assist with drafting and revising manuscript text, grammar and style editing, literature search and citation checking, discussion of experimental results, reviewer-style feedback, and figure design. We also used Codex CLI with GPT-6-Astra at the high reasoning-effort setting to discuss and implement the proposed method, baseline methods, and experimental code.

The experimental VLMs, including Qwen3-VL, Ministral 3, Gemma 4, InternVL3.5, and MiniCPM-V4.5, were used for slot ranking and environment-label inference, with additional Qwen3-VL scoring and direct-prediction experiments (Sec.~\ref{sec:experiments}, Apps.~\ref{app:detailed_results} and~\ref{app:vlm_free_inference}). GPT-6-Astra also provided task-evidence labels for the independent annotation analysis in App.~\ref{app:ranking_evidence}. The authors retain full responsibility for the scientific claims, experimental results, code, and references.

\section*{Acknowledgments}
This material is based upon work supported by National Science and Technology Council, ROC under grant number 114-2221-E-002-134-MY3, NTU AI Center of Research Excellence within Taiwan Centers of Excellence in Artificial Intelligence, and by National Taiwan University and Academia Sinica Innovative Joint Program, under grant AS-NTU-114-06.
{
    \small
    \bibliographystyle{ieeenat_fullname}
    \bibliography{main}
}

% Compile with the appendix to preserve cross-references, then split the PDF.
% The main submission contains the main text, LLM Usage Statement, and references.
% Upload the appendix pages separately as supplementary material.
% \input{sec/X_suppl}
\clearpage
\onecolumn
\appendix
\raggedbottom

% The supplementary material uses one column throughout.
% Large study blocks begin on a fresh page; all tables remain in text order.
\newenvironment{appwidetable}{%
  \par\medskip\noindent\begin{minipage}{\textwidth}%
  \centering\captionsetup{type=table}%
}{\end{minipage}\par\medskip}
\newenvironment{appwidefigure}{%
  \par\medskip\noindent\begin{minipage}{\textwidth}%
  \centering\captionsetup{type=figure}%
}{\end{minipage}\par\medskip}
\newenvironment{appcompacttable}{%
  \par\medskip\begingroup\centering
  \begin{minipage}{0.72\textwidth}%
  \centering\captionsetup{type=table}%
}{\end{minipage}\par\endgroup\medskip}

\clearpage

\section{Experimental Details \& Numerical Results}
\label{sec:appendix}

Bold labels identify default design choices. Bold accuracies and gains mark the highest mean in each comparison.

\subsection{Datasets and Data Preparation}
\label{app:datasets}

Table~\ref{tab:dataset_setup} lists the image counts after deduplication.

\begin{appwidetable}
\caption{Dataset splits and image counts after deduplication. Each NICO++ row names the two environments held out exclusively for testing. The remaining four environments supply training and validation data.}
\label{tab:dataset_setup}
\centering
\small
\setlength{\tabcolsep}{4pt}
\renewcommand{\arraystretch}{1.05}
\begin{tabularx}{\textwidth}{@{}>{\raggedright\arraybackslash}p{0.18\textwidth}>{\raggedright\arraybackslash}Xrrr>{\raggedright\arraybackslash}p{0.15\textwidth}@{}}
\toprule
Dataset / split & Prediction task and environment & Train & Val. & Test & Test groups \\
\midrule
Waterbirds & Landbird/waterbird. Land/water background. & 4,795 & 1,199 & 5,794 & Class $\times$ environment (4) \\
CelebA & Binary \texttt{Blond\_Hair}. Grouping attribute: \texttt{Male}. & 162,647 & 19,861 & 19,954 & Class $\times$ environment (4) \\
MetaShift & Cat/dog. Indoor/outdoor context. & 1,360 & 340 & 576 & Class $\times$ environment (4) \\
NICO++:\newline Autumn/Rock & 60 classes. Six environments. & 53,589 & 13,399 & 19,730 & Target environments (2) \\
NICO++:\newline Dim/Grass & Same task and environment pool. & 43,768 & 10,943 & 32,007 & Target environments (2) \\
NICO++:\newline Outdoor/Water & Same task and environment pool. & 41,389 & 10,348 & 34,981 & Target environments (2) \\
\bottomrule
\end{tabularx}

\end{appwidetable}

\paragraph{Data partitions.}
Waterbirds uses its official split. CelebA uses the deduplicated official split, preserving the original split assignment of each retained image. For MetaShift, we use the official default subpopulation-shift setting, reserve 20\% of its original training set for validation by stratifying on class $\times$ environment, and use the original held-out set for testing. For each NICO++ split, the two target environments are used exclusively for testing. Each of the other four source environments is split 80\%/20\% into training/validation data, stratified by class. Custom MetaShift and NICO++ partitions use split seed 42. A NICO++ environment reuses the same training/validation lists whenever it serves as a source in different splits.

\paragraph{Deduplication.}
For CelebA and NICO++, we retain one copy of each exact duplicate image file. If copies have conflicting class or environment labels, we discard the entire duplicate cluster. CelebA prioritizes test over validation over training and retains the first image ID within the selected split, reducing its total from 202,599 to 202,462. For NICO++, exact-duplicate copies are resolved in image-ID order before splitting, reducing the evaluated six-environment pool from 88,866 to 86,718. Cropped, recompressed, and other near-duplicate images are not merged.

\subsection{Evaluation and Model Selection}
\label{app:evaluation}

\paragraph{Test groups.}
Ground-truth labels define four class $\times$ environment groups for Waterbirds, CelebA, and MetaShift. NICO++ uses the two held-out environments as test groups to avoid small class $\times$ environment cells. Within each NICO++ environment, accuracy is computed over images without additional class balancing.

\paragraph{Aggregation and reporting.}
For each training seed, we average the three NICO++ splits and then weight the four datasets equally, separately for average-group and worst-group accuracy. Aggregation precedes the across-seed summary, so macro-averaged worst-group accuracy aggregates within-split worst-group results.

\paragraph{Model selection.}
All predictor experiments use the same method-specific hyperparameter search spaces. For each method, split, and condition, candidates use a common set of training seeds. Each seed uses the sum of validation balanced accuracy and macro F1 for model selection, with zero-denominator F1 values set to zero. For SWAD and DFR, evaluation uses the parameter-averaged predictor. We select one candidate by its mean validation score across seeds and use it for the corresponding test runs. Ties follow the fixed candidate enumeration order. Each NICO++ split and each comparison condition is selected independently. Test performance is never used for selection.

\subsection{Representation and Training Implementation}
\label{app:implementation}

\paragraph{Frozen representations and ranking.}
Unless varied in an ablation, we use the frozen ViT-S/14 FT-DINOSAUR checkpoint with seven slots, top-3 decoding, and fixed slot initialization. Images are resized to $224\times224$ with bicubic interpolation and normalized with ImageNet statistics. The seven 256-dimensional slot embeddings are concatenated into 1,792 dimensions. Hard masks are obtained by argmax on the $16\times16$ patch grid and upsampled with nearest-neighbor interpolation. Each slot image retains its assigned region and fills the remaining pixels with RGB $(128,128,128)$. Slots receiving no patch are empty.
% Checkpoint: dinosaur_small_patch14_224_topk3.coco_dv2_ft_s7_300k-fbc7a7c7.pth.

Frozen Qwen3-VL-8B-Instruct implements Direct Rank with greedy constrained decoding. Nonempty slot images are identified by their original slot IDs, and the output must contain each ID exactly once. Empty slots are appended in a deterministic per-image random order, fixed across runs. Cached rankings are shared across training seeds. App.~\ref{app:vlm_prompts} lists all VLM checkpoint identifiers, complete prompt templates, dataset-specific task descriptions, and evidence criteria.

\paragraph{Preprocessing and predictor.}
Unless otherwise specified, each input coordinate is standardized using the training mean and standard deviation for the corresponding slot order, then fixed for validation and test. Zero-variance coordinates use a divisor of one. For ranked inputs, this implements $T_r$ in Eq.~\eqref{eq:ranked_slots}. The default predictor is Linear--BatchNorm--ReLU--Linear with 64 hidden units and $C$ output logits. Hard removal changes its input dimension but not its hidden width.

\paragraph{Loss weighting and sampling.}
CE, RSC, Mixup, Rank2Replace, Grouped LAAT, and hard removal average CE within each class present in the batch, then equally across these classes. IRM and Group DRO use group-balanced sampling: class $\times$ environment groups for the first three datasets and the four source environments for NICO++. CORAL balances environments only, and JTT uses error-based resampling. These four methods use ordinary CE without additional class weights.
ElRep, SAGM, SWAD, and SCER use natural sampling and class-balanced CE. DFR balances class $\times$ environment groups during head refitting on all four datasets (App.~\ref{app:baselines}).

\paragraph{Optimization.}
The shared predictor training stage uses Adam with learning rate $10^{-4}$, weight decay 0, and batch size 256. The extractor remains frozen unless varied in an ablation. The default training seeds are 9803, 38224, and 8113. Runs last at most 5,000 epochs and stop after 200 epochs without strict validation-score improvement, retaining the earliest best checkpoint when applicable. JTT's first stage additionally runs for the candidate-specific $T$ epochs, with the common budget applied to its second stage. Unless specified otherwise, validation and test use uncorrupted inputs without Mixup or RSC masking, while retaining the method's slot order and hard removal.

\subsection{Baseline Implementations and Search Spaces}
\label{app:baselines}

Table~\ref{tab:baseline_search} lists method-specific search spaces for $K=7$. Implementation details follow.

\begin{appwidetable}
\caption{Method-specific search spaces for the default $K=7$ configuration. Slot order refers to Overall Performance (Sec.~\ref{sec:overall_performance}).}
\label{tab:baseline_search}
\centering
\small
\setlength{\tabcolsep}{5pt}
\renewcommand{\arraystretch}{1.05}
\begin{tabularx}{\textwidth}{@{}p{0.19\textwidth}p{0.12\textwidth}X@{}}
\toprule
Method & Slot order & Method-specific hyperparameter space \\
\midrule
CE & Original & No method-specific hyperparameters. \\
IRM & Original & $\lambda\in\{0.003,0.01,0.1,1,10,100,1000,10000\}$. \\
Group DRO & Original & $\eta\in\{0.001,0.01,0.1,1,3,10\}$. \\
JTT & Original & $T\in\{1,5,10,50,100,200\}$, $u\in\{1.5,2,5,20,100,300\}$. Full $6\times6$ grid. \\
RSC & Original & $p_f\in\{0.05,0.1,1/3,0.5,0.75,0.9\}$, $p_b\in\{0.02,0.05,0.1,1/3,0.5\}$. Full $6\times5$ grid. \\
CORAL & Original & $\lambda\in\{0.003,0.01,0.1,1,10,100,300\}$. \\
Mixup & Original & $\alpha\in\{0.03,0.1,0.2,0.5,1,2,5\}$. \\
ElRep & Original & $\lambda_{\mathrm{N}}\in\{10^{-4},3\!\times\!10^{-4},10^{-3}\}$, $\lambda_{\mathrm{F}}\in\{10^{-5},3\!\times\!10^{-5},10^{-4}\}$. Full $3\times3$ grid. \\
SAGM & Original & $\alpha\in\{0.0005,0.001\}$. \\
SWAD & Original & Tolerance $\tau\in\{0.2,0.3\}$. \\
SCER & Original & $\lambda_{\mathrm{spur}},\lambda_{\mathrm{core}}\in\{0.1,0.5,1\}$. Full $3\times3$ grid. \\
DFR & Original & Fixed balanced head-refitting protocol. No method-specific search. \\
\textbf{Rank2Replace} & \textbf{Ranked} & Default Top-preserving schedule in Sec.~\ref{sec:slot_corruption}. No method-specific search. \\
Grouped LAAT & \textbf{Ranked} & $\lambda_{\mathrm{A}}\in\{0.001,0.01,0.03,0.1,0.5,1,2,5,10,100,1000\}$. \\
Hard removal & \textbf{Ranked} & $k\in\{1,2,3,4,5,6\}$. \\
\bottomrule
\end{tabularx}

\end{appwidetable}

\paragraph{CE, IRM, and Group DRO.}
CE uses natural sampling with class-balanced CE. IRM adds its invariance penalty to the group-balanced objective. The penalty coefficient is one for the first 500 optimizer updates, after which it switches to the selected $\lambda$ and Adam is reset. Group DRO dynamically updates group weights with step size $\eta$, uses ordinary CE within groups, and fixes generalization adjustment to zero.

\paragraph{JTT and RSC.}
JTT first trains with ordinary CE for exactly $T$ epochs and fixes the error set at that epoch. Its second stage reinitializes the predictor and uses weighted sampling with replacement, assigning weight $u$ to the misclassified examples, with ordinary CE. RSC applies feature and sample masking to the 64-dimensional hidden representation, controlled by $p_f$ and $p_b$, while retaining class-balanced CE.

\paragraph{CORAL.}
CORAL uses ordinary CE plus alignment of hidden means and unbiased covariance estimates across environments, weighted by $\lambda$. Squared differences are averaged over all environment pairs. Alignment acts on the 64-dimensional activations after BatchNorm and ReLU, using one BatchNorm forward pass on the combined batch. Both sampling and loss aggregation give equal weight to environments rather than class $\times$ environment groups.

\paragraph{Mixup.}
Mixup uses natural sampling and mixes the full standardized, concatenated slot embeddings. Each batch draws one coefficient $\rho\sim\operatorname{Beta}(\alpha,\alpha)$ and a uniform pairing permutation, allowing self-pairs. The same coefficient weights the two class-balanced CE terms for the paired labels.

\paragraph{ElRep and SCER.}
Both regularizers act on the 64-dimensional hidden activations after BatchNorm and ReLU. ElRep adds nuclear- and squared Frobenius-norm penalties, each divided by the batch size and weighted by $\lambda_{\mathrm{N}}$ and $\lambda_{\mathrm{F}}$. SCER estimates spurious and core directions from minibatch class--environment mean activations, omitting empty cells. It penalizes spurious-direction alignment and encourages core-direction alignment, with prediction-head weights detached within the regularizer. SCER uses Qwen3-VL environment labels unless the label source is varied.

\paragraph{SAGM and SWAD.}
SAGM uses the nonadaptive update with fixed perturbation radius $\rho=0.05$ and the shared Adam optimizer. SWAD averages weights after every optimizer update and detects the averaging interval from validation class-balanced CE on uncorrupted inputs, using convergence and tolerance windows of three and six epochs. Its loss threshold is $(1+\tau)$ times the mean loss in the convergence window. BatchNorm statistics are recalibrated on training inputs before evaluating the final averaged predictor with the common validation selection score.

\paragraph{DFR.}
Our DFR adaptation trains the MLP on 80\% of the training set, freezes its hidden representation, and refits the prediction head on the remaining 20\%. Refitting uses class $\times$ environment-balanced sampling, with Qwen3-VL environment labels unless their source is varied. Head refitting follows the official DFR procedure, averaging the parameters of ten refitted heads without an additional method-specific search. Both stages use training images only.

\paragraph{Grouped LAAT.}
Grouped LAAT is a slot-level adaptation of Large Language Model Attribution Aligned Training. For ranked slot embeddings $z_i$, we define a ground-truth logit margin and slot importance,
\begin{equation}
\begin{aligned}
\Delta_i
  &= [f_\theta(z_i)]_{y_i}
     -\frac{1}{C-1}\sum_{c\ne y_i}[f_\theta(z_i)]_c,\\
a_{i,r}
  &= \left\|z_{i,r}\odot
        \nabla_{z_{i,r}}\Delta_i\right\|_2.
\end{aligned}
\label{eq:grouped_laat_importance}
\end{equation}
The seven-dimensional vector $a_i=(a_{i,1},\ldots,a_{i,7})$ and the prior $(7,6,5,4,3,2,1)$ are separately $\ell_2$ normalized and compared by mean squared error. We add this alignment term to class-balanced CE with coefficient $\lambda_{\mathrm{A}}$. Importance is computed in evaluation mode, and CE in training mode. The constraint acts across the seven slot groups without prescribing the attribution distribution among the 256 coordinates within a slot. The prior comes from slot ranking without additional scoring.

\paragraph{Hard removal.}
Hard removal retains only the top $k$ ranked slots and trains with class-balanced CE. Its search always removes at least one slot. In the slot-count ablation, $k$ is selected separately for each $K$ using validation data.

\clearpage
\subsection{Detailed Results and Ablation Settings}
\label{app:detailed_results}

The subsections follow the order of Secs.~\ref{sec:main_results} and~\ref{sec:ablations}. They supply additional settings and numerical results. Tables already shown in the main paper are referenced rather than repeated.

\subsubsection{Global Embeddings and End-to-End Training}
\label{app:global_image_embeddings}

Table~\ref{tab:global_macro} in Sec.~\ref{sec:global_image_embeddings} reports the complete macro comparison. The following settings supplement the shared protocol.

\paragraph{Representation.}
Using the same FT-DINOSAUR ViT-S/14 encoder checkpoint, we mean-pool the 256 patch embeddings of each $224\times224$ image into a 384-dimensional global image embedding, without Slot Attention, the decoder, or VLM ranking. Frozen conditions train a linear predictor or the shared MLP. End-to-end CE jointly trains the encoder and linear predictor with class-balanced CE.

End-to-end Rank2Replace jointly updates the encoder, Slot Attention, and MLP predictor. Qwen3-VL remains frozen, and cached rankings, slot initialization, and normalization statistics remain fixed. The original segmentation masks are used only to obtain rankings, not during training. Slot embeddings and attention assignments evolve with training, but segmentation masks and rankings are not regenerated.

\paragraph{AFR and SagNet.}
AFR~\cite{pmlr-v202-qiu23c} uses a class-stratified, disjoint 80\%/20\% training partition. It trains the encoder and linear predictor on the former with class-balanced CE, then freezes the encoder and refits the head from its first-stage parameters on the latter. Refitting uses weighted CE with fixed, normalized weights $w_i\propto n_{y_i}^{-1}\exp(-\gamma p_i)$, where $p_i$ is the first-stage true-class probability and $n_{y_i}$ is the class count in the refitting subset. Our ViT adaptation of SagNet~\cite{Nam2021ReducingDG} branches after block six of twelve, with separate remaining blocks and linear predictors. The content branch randomizes style statistics, and the style branch replaces normalized content. Both use class-balanced CE. An adversarial update encourages uniform style predictions by updating only the shared front-end LayerNorm affine parameters. Statistics use patch tokens only, but the classification token receives the same transformation. Donor pairing spans the full batch with gradients preserved across microbatches. Inference retains only the main encoder, mean pooling, and linear predictor.

\paragraph{Training.}
Frozen predictors use App.~\ref{app:implementation}. End-to-end CE, AFR's first stage, and SagNet's classification updates use Adam with encoder learning rate $\in\{10^{-5},10^{-4}\}$, predictor learning rate $10^{-4}$, batch size 256 (microbatches of 32), and at most 50 epochs with patience 10. AFR refitting searches $\gamma\in\{1,2,4,8\}$ and uses full-batch SGD with learning rate 0.01, zero momentum, gradient clipping 1, and at most 1,000 steps with patience 200. SagNet's adversarial updates use SGD with learning rate 0.004, momentum 0.9, gradient clipping 0.1, and $\lambda_{\mathrm{adv}}\in\{0.01,0.1,1\}$. The encoder learning rate and each method's additional coefficient are selected jointly. For these global-image conditions, all stages use zero weight decay, and AFR adds no head regularization. End-to-end CE and AFR's first stage use BF16 encoder autocasting with FP32 parameters and predictors. SagNet uses BF16 encoder computation with FP32 style statistics and predictors. Global image embeddings use fixed training-only standardization fitted on the full training set, or on the first-stage 80\% for AFR and reused during refitting. Global-image CE and SagNet use the full training set. The global-image conditions use no additional image augmentation or environment labels.

\clearpage
\subsubsection{Overall Performance}
\label{app:overall_results}

Tables~\ref{tab:overall_context} and~\ref{tab:overall_nico} report the split-level results in Fig.~\ref{fig:baseline_performance}. Macro results are printed in the figure.

\begin{appwidetable}
\caption{Overall performance in Fig.~\ref{fig:baseline_performance}: Waterbirds, CelebA, and MetaShift.}
\label{tab:overall_context}
\centering
\small
\setlength{\tabcolsep}{3pt}
\renewcommand{\arraystretch}{1.04}
\begin{tabular*}{\textwidth}{@{\extracolsep{\fill}}lrrrrrr@{}}
\toprule
 & \multicolumn{2}{c}{Waterbirds} & \multicolumn{2}{c}{CelebA} & \multicolumn{2}{c}{MetaShift} \\
\cmidrule(lr){2-3}\cmidrule(lr){4-5}\cmidrule(lr){6-7}
Method & Average-group & Worst-group & Average-group & Worst-group & Average-group & Worst-group \\
\midrule
CE & $68.87 \pm 0.55$ & $37.12 \pm 2.68$ & $82.73 \pm 1.27$ & $51.67 \pm 4.34$ & $73.38 \pm 2.12$ & $43.52 \pm 7.43$ \\
IRM & $75.33 \pm 1.59$ & $60.45 \pm 8.93$ & $83.43 \pm 0.79$ & $55.93 \pm 3.21$ & $76.97 \pm 0.53$ & $56.71 \pm 4.19$ \\
Group DRO & $75.55 \pm 1.14$ & $54.00 \pm 7.09$ & $83.83 \pm 2.15$ & $56.30 \pm 9.17$ & $74.88 \pm 1.60$ & $45.14 \pm 8.67$ \\
JTT & $73.41 \pm 1.17$ & $49.90 \pm 4.56$ & $76.62 \pm 0.91$ & $28.33 \pm 4.01$ & $73.78 \pm 1.84$ & $45.83 \pm 5.56$ \\
RSC & $69.99 \pm 0.33$ & $35.05 \pm 1.80$ & $82.24 \pm 0.42$ & $48.33 \pm 1.47$ & $74.36 \pm 2.16$ & $45.83 \pm 9.34$ \\
CORAL & $72.76 \pm 0.73$ & $60.38 \pm 3.71$ & $76.21 \pm 0.48$ & $27.59 \pm 2.10$ & $72.80 \pm 0.87$ & $40.51 \pm 4.01$ \\
Mixup & $69.14 \pm 0.99$ & $37.75 \pm 1.64$ & $82.93 \pm 1.10$ & $51.85 \pm 4.46$ & $74.13 \pm 2.11$ & $44.21 \pm 9.06$ \\
ElRep & $69.13 \pm 0.38$ & $36.40 \pm 0.48$ & $82.98 \pm 0.59$ & $52.22 \pm 2.94$ & $73.38 \pm 2.24$ & $43.29 \pm 7.36$ \\
SAGM & $69.60 \pm 0.83$ & $38.63 \pm 0.27$ & $83.11 \pm 0.79$ & $51.67 \pm 3.38$ & $73.32 \pm 1.65$ & $42.59 \pm 6.30$ \\
SWAD & $65.85 \pm 1.06$ & $28.56 \pm 2.75$ & $83.47 \pm 0.30$ & $53.52 \pm 1.16$ & $74.59 \pm 0.96$ & $49.31 \pm 2.78$ \\
SCER & $69.04 \pm 0.26$ & $33.54 \pm 0.39$ & $82.96 \pm 0.28$ & $51.48 \pm 1.70$ & $74.48 \pm 1.14$ & $45.60 \pm 6.45$ \\
DFR & $66.77 \pm 3.11$ & $46.26 \pm 2.97$ & $\mathbf{88.13} \pm 0.18$ & $\mathbf{84.05} \pm 0.62$ & $70.43 \pm 1.82$ & $52.55 \pm 5.12$ \\
\textbf{Rank2Replace} & $\mathbf{82.49} \pm 0.84$ & $\mathbf{63.08} \pm 0.62$ & $85.13 \pm 0.45$ & $60.93 \pm 1.16$ & $\mathbf{83.33} \pm 0.76$ & $68.75 \pm 1.84$ \\
Grouped LAAT & $74.09 \pm 1.04$ & $50.42 \pm 1.96$ & $82.16 \pm 0.74$ & $48.33 \pm 2.78$ & $80.44 \pm 1.13$ & $60.42 \pm 4.22$ \\
Hard removal & $81.02 \pm 0.46$ & $59.50 \pm 1.08$ & $83.72 \pm 0.23$ & $53.33 \pm 0.96$ & $82.99 \pm 0.35$ & $\mathbf{68.98} \pm 0.40$ \\
\bottomrule
\end{tabular*}

\end{appwidetable}
\begin{appwidetable}
\caption{Overall performance in Fig.~\ref{fig:baseline_performance}: the three NICO++ splits.}
\label{tab:overall_nico}
\centering
\small
\setlength{\tabcolsep}{3pt}
\renewcommand{\arraystretch}{1.04}
\begin{tabular*}{\textwidth}{@{\extracolsep{\fill}}lrrrrrr@{}}
\toprule
 & \multicolumn{2}{c}{Autumn/Rock} & \multicolumn{2}{c}{Dim/Grass} & \multicolumn{2}{c}{Outdoor/Water} \\
\cmidrule(lr){2-3}\cmidrule(lr){4-5}\cmidrule(lr){6-7}
Method & Average-group & Worst-group & Average-group & Worst-group & Average-group & Worst-group \\
\midrule
CE & $64.09 \pm 0.32$ & $63.30 \pm 0.64$ & $62.83 \pm 0.39$ & $59.81 \pm 0.48$ & $58.84 \pm 0.26$ & $53.15 \pm 0.45$ \\
IRM & $63.16 \pm 0.41$ & $62.30 \pm 0.26$ & $61.65 \pm 0.14$ & $58.22 \pm 0.40$ & $56.92 \pm 0.36$ & $51.91 \pm 0.36$ \\
Group DRO & $62.73 \pm 0.03$ & $62.01 \pm 0.10$ & $62.55 \pm 0.54$ & $59.33 \pm 0.60$ & $58.18 \pm 0.18$ & $53.07 \pm 0.43$ \\
JTT & $63.48 \pm 0.22$ & $62.92 \pm 0.28$ & $62.31 \pm 0.63$ & $58.88 \pm 0.84$ & $58.53 \pm 0.35$ & $53.50 \pm 0.26$ \\
RSC & $64.75 \pm 0.11$ & $63.87 \pm 0.06$ & $63.07 \pm 0.49$ & $60.38 \pm 0.69$ & $59.11 \pm 0.15$ & $53.81 \pm 0.33$ \\
CORAL & $62.78 \pm 0.52$ & $61.68 \pm 0.45$ & $61.85 \pm 0.05$ & $58.51 \pm 0.11$ & $57.24 \pm 0.36$ & $52.75 \pm 0.36$ \\
Mixup & $62.97 \pm 0.39$ & $62.39 \pm 0.27$ & $61.27 \pm 0.38$ & $58.65 \pm 0.13$ & $58.62 \pm 0.46$ & $52.95 \pm 0.24$ \\
ElRep & $64.29 \pm 0.57$ & $63.44 \pm 0.74$ & $62.92 \pm 0.58$ & $59.83 \pm 0.55$ & $58.84 \pm 0.26$ & $53.15 \pm 0.42$ \\
SAGM & $64.52 \pm 0.04$ & $63.57 \pm 0.20$ & $63.24 \pm 0.30$ & $60.28 \pm 0.21$ & $58.83 \pm 0.21$ & $53.23 \pm 0.36$ \\
SWAD & $64.36 \pm 0.25$ & $63.61 \pm 0.57$ & $62.46 \pm 0.43$ & $59.39 \pm 0.45$ & $58.54 \pm 0.24$ & $53.14 \pm 0.43$ \\
SCER & $64.23 \pm 0.59$ & $63.41 \pm 0.70$ & $62.85 \pm 0.27$ & $59.95 \pm 0.43$ & $58.88 \pm 0.28$ & $53.20 \pm 0.51$ \\
DFR & $56.20 \pm 0.70$ & $56.01 \pm 0.63$ & $56.46 \pm 1.44$ & $53.19 \pm 1.91$ & $52.83 \pm 0.73$ & $47.02 \pm 0.56$ \\
\textbf{Rank2Replace} & $\mathbf{72.49} \pm 0.04$ & $\mathbf{72.14} \pm 0.13$ & $\mathbf{71.69} \pm 0.09$ & $\mathbf{68.57} \pm 0.12$ & $\mathbf{68.15} \pm 0.18$ & $\mathbf{64.33} \pm 0.29$ \\
Grouped LAAT & $70.83 \pm 0.56$ & $70.10 \pm 0.60$ & $69.17 \pm 0.29$ & $66.23 \pm 0.49$ & $65.36 \pm 0.29$ & $60.72 \pm 0.16$ \\
Hard removal & $72.30 \pm 0.30$ & $71.60 \pm 0.19$ & $71.56 \pm 0.39$ & $68.33 \pm 0.44$ & $67.58 \pm 0.53$ & $63.53 \pm 0.33$ \\
\bottomrule
\end{tabular*}

\end{appwidetable}

\clearpage
\subsubsection{Contributions and Compatibility}
\label{app:ranking_compatibility}

Tables~\ref{tab:compatibility_average} and~\ref{tab:compatibility_worst} provide the results in Fig.~\ref{fig:compatibility_macro}. Ranking gain compares slot ranking with the original slot order. Corruption gain compares ranked inputs with and without slot corruption, and total gain compares ranking plus corruption with the original slot order.

\begin{appwidetable}
\caption{Average-group accuracy and gains for Fig.~\ref{fig:compatibility_macro}.}
\label{tab:compatibility_average}
\centering
\small
\setlength{\tabcolsep}{5pt}
\renewcommand{\arraystretch}{1.04}
\begin{tabular}{@{}lrrrrrr@{}}
\toprule
Method & \shortstack{Original\\slot order} & \shortstack{Slot\\ranking} & \textbf{\shortstack{Ranking +\\corruption}} & \shortstack{Ranking\\gain} & \shortstack{Corruption\\gain} & \shortstack{Total\\gain} \\
\midrule
\textbf{CE} & $71.72 \pm 0.66$ & $75.96 \pm 0.29$ & $80.43 \pm 0.33$ & $+4.24$ & $+4.47$ & $+8.71$ \\
IRM & $\mathbf{74.07} \pm 0.27$ & $77.55 \pm 0.56$ & $\mathbf{81.96} \pm 0.08$ & $+3.48$ & $+4.41$ & $+7.89$ \\
Group DRO & $73.85 \pm 0.36$ & $77.42 \pm 0.55$ & $81.38 \pm 0.23$ & $+3.57$ & $+3.96$ & $+7.53$ \\
JTT & $71.31 \pm 0.62$ & $75.35 \pm 0.71$ & $79.18 \pm 0.33$ & $+4.04$ & $+3.83$ & $+7.87$ \\
RSC & $72.23 \pm 0.65$ & $76.03 \pm 0.14$ & $80.19 \pm 0.48$ & $+3.80$ & $+4.16$ & $+7.96$ \\
CORAL & $70.60 \pm 0.23$ & $75.15 \pm 0.07$ & $78.82 \pm 0.50$ & $\mathbf{+4.55}$ & $+3.67$ & $+8.22$ \\
Mixup & $71.79 \pm 0.77$ & $75.79 \pm 0.20$ & $80.50 \pm 0.29$ & $+4.00$ & $+4.71$ & $+8.71$ \\
ElRep & $71.88 \pm 0.56$ & $76.05 \pm 0.78$ & $80.42 \pm 0.39$ & $+4.17$ & $+4.37$ & $+8.54$ \\
SAGM & $72.06 \pm 0.19$ & $76.14 \pm 0.73$ & $80.53 \pm 0.17$ & $+4.08$ & $+4.39$ & $+8.47$ \\
SWAD & $71.43 \pm 0.08$ & $75.43 \pm 0.11$ & $80.25 \pm 0.09$ & $+4.00$ & $\mathbf{+4.82}$ & $\mathbf{+8.82}$ \\
SCER & $72.12 \pm 0.35$ & $75.77 \pm 0.23$ & $80.44 \pm 0.15$ & $+3.65$ & $+4.67$ & $+8.32$ \\
DFR & $70.12 \pm 0.42$ & $74.62 \pm 0.20$ & $78.42 \pm 0.12$ & $+4.50$ & $+3.80$ & $+8.30$ \\
Grouped LAAT & -- & $76.28 \pm 0.58$ & $80.39 \pm 0.14$ & -- & $+4.11$ & -- \\
Hard removal & -- & $\mathbf{79.55} \pm 0.07$ & $80.50 \pm 0.24$ & -- & $+0.95$ & -- \\
\bottomrule
\end{tabular}

\end{appwidetable}
\begin{appwidetable}
\caption{Worst-group accuracy and gains for Fig.~\ref{fig:compatibility_macro}.}
\label{tab:compatibility_worst}
\centering
\small
\setlength{\tabcolsep}{5pt}
\renewcommand{\arraystretch}{1.04}
\begin{tabular}{@{}lrrrrrr@{}}
\toprule
Method & \shortstack{Original\\slot order} & \shortstack{Slot\\ranking} & \textbf{\shortstack{Ranking +\\corruption}} & \shortstack{Ranking\\gain} & \shortstack{Corruption\\gain} & \shortstack{Total\\gain} \\
\midrule
\textbf{CE} & $47.77 \pm 3.02$ & $55.48 \pm 0.63$ & $65.28 \pm 0.81$ & $+7.71$ & $+9.80$ & $+17.51$ \\
IRM & $57.64 \pm 2.79$ & $62.59 \pm 1.24$ & $\mathbf{72.68} \pm 0.67$ & $+4.95$ & $+10.09$ & $+15.04$ \\
Group DRO & $53.39 \pm 2.87$ & $61.02 \pm 2.70$ & $68.03 \pm 0.86$ & $+7.63$ & $+7.01$ & $+14.64$ \\
JTT & $45.62 \pm 1.69$ & $51.53 \pm 4.29$ & $59.80 \pm 1.26$ & $+5.91$ & $+8.27$ & $+14.18$ \\
RSC & $47.14 \pm 2.03$ & $53.95 \pm 1.32$ & $65.05 \pm 1.25$ & $+6.81$ & $+11.10$ & $+17.91$ \\
CORAL & $46.53 \pm 0.75$ & $54.10 \pm 1.69$ & $60.25 \pm 0.48$ & $+7.57$ & $+6.15$ & $+13.72$ \\
Mixup & $47.95 \pm 3.07$ & $55.59 \pm 1.18$ & $65.72 \pm 0.92$ & $+7.64$ & $+10.13$ & $+17.77$ \\
ElRep & $47.68 \pm 2.16$ & $56.06 \pm 2.12$ & $65.14 \pm 1.50$ & $\mathbf{+8.38}$ & $+9.08$ & $+17.46$ \\
SAGM & $47.98 \pm 1.89$ & $55.85 \pm 1.45$ & $65.49 \pm 0.73$ & $+7.87$ & $+9.64$ & $+17.51$ \\
SWAD & $47.52 \pm 0.54$ & $53.98 \pm 0.38$ & $64.44 \pm 0.23$ & $+6.46$ & $+10.46$ & $+16.92$ \\
SCER & $47.37 \pm 1.95$ & $54.05 \pm 1.17$ & $65.31 \pm 0.59$ & $+6.68$ & $\mathbf{+11.26}$ & $\mathbf{+17.94}$ \\
DFR & $\mathbf{58.73} \pm 1.26$ & $\mathbf{63.29} \pm 1.34$ & $69.57 \pm 1.64$ & $+4.56$ & $+6.28$ & $+10.84$ \\
Grouped LAAT & -- & $56.21 \pm 1.49$ & $65.16 \pm 0.53$ & -- & $+8.95$ & -- \\
Hard removal & -- & $62.41 \pm 0.32$ & $65.55 \pm 0.56$ & -- & $+3.14$ & -- \\
\bottomrule
\end{tabular}

\end{appwidetable}

\clearpage
\subsubsection{Environment-Label Quality}
\label{app:environment_quality}

Table~\ref{tab:environment_quality} reports the conditions in Fig.~\ref{fig:environment_downstream}. Only training environment assignments vary. Class labels and ground-truth evaluation groups remain unchanged. Cached assignments are shared across methods and hyperparameter candidates for each split, label source, and training seed. App.~\ref{app:environment_prompts} gives the annotation prompts and allowed labels.

\begin{appwidetable}
\caption{Environment-label conditions in Fig.~\ref{fig:environment_downstream}. Bold accuracies mark each baseline's best label source.}
\label{tab:environment_quality}
\centering
\small
\setlength{\tabcolsep}{2pt}
\renewcommand{\arraystretch}{1.04}
\begin{tabular}{@{}llrrrrr@{}}
\toprule
Environment labels & Metric & IRM & Group DRO & CORAL & SCER & DFR \\
\midrule
Oracle & Average-group & $\mathbf{75.29} \pm 0.30$ & $74.42 \pm 0.67$ & $\mathbf{71.71} \pm 0.34$ & $\mathbf{72.76} \pm 0.15$ & $71.33 \pm 0.84$ \\
 & Worst-group & $\mathbf{61.27} \pm 1.89$ & $55.66 \pm 2.92$ & $47.25 \pm 1.19$ & $\mathbf{49.39} \pm 1.11$ & $\mathbf{62.50} \pm 1.72$ \\
\addlinespace[2pt]
Oracle + 10\% noise & Average-group & $73.25 \pm 0.14$ & $73.92 \pm 0.90$ & $70.42 \pm 0.85$ & $71.94 \pm 0.70$ & $70.46 \pm 0.50$ \\
 & Worst-group & $54.29 \pm 1.55$ & $53.70 \pm 4.29$ & $42.10 \pm 2.45$ & $48.27 \pm 1.54$ & $54.55 \pm 1.99$ \\
\addlinespace[2pt]
Oracle + 20\% noise & Average-group & $71.99 \pm 0.76$ & $73.09 \pm 0.52$ & $69.70 \pm 0.27$ & $71.39 \pm 0.57$ & $70.25 \pm 0.34$ \\
 & Worst-group & $50.13 \pm 3.09$ & $50.85 \pm 2.18$ & $40.42 \pm 0.84$ & $46.44 \pm 2.60$ & $51.67 \pm 0.46$ \\
\addlinespace[2pt]
Oracle + 30\% noise & Average-group & $71.46 \pm 0.32$ & $71.92 \pm 0.17$ & $70.00 \pm 0.43$ & $71.22 \pm 0.57$ & $70.06 \pm 0.21$ \\
 & Worst-group & $46.17 \pm 2.43$ & $48.11 \pm 2.94$ & $40.17 \pm 0.23$ & $44.57 \pm 2.27$ & $50.59 \pm 1.37$ \\
\addlinespace[2pt]
Oracle + 40\% noise & Average-group & $71.00 \pm 0.37$ & $70.90 \pm 0.62$ & $69.73 \pm 0.22$ & $71.65 \pm 0.58$ & $69.79 \pm 0.32$ \\
 & Worst-group & $44.45 \pm 1.66$ & $43.54 \pm 2.70$ & $39.18 \pm 0.48$ & $47.40 \pm 2.53$ & $50.60 \pm 0.72$ \\
\addlinespace[2pt]
Oracle + 50\% noise & Average-group & $70.65 \pm 0.22$ & $71.20 \pm 0.63$ & $69.97 \pm 0.63$ & $71.38 \pm 0.36$ & $69.60 \pm 0.56$ \\
 & Worst-group & $42.70 \pm 0.75$ & $44.14 \pm 1.54$ & $38.31 \pm 2.07$ & $45.77 \pm 1.78$ & $50.45 \pm 0.72$ \\
\midrule
\textbf{Qwen3-VL} & Average-group & $74.07 \pm 0.27$ & $73.85 \pm 0.36$ & $70.60 \pm 0.23$ & $72.12 \pm 0.35$ & $70.12 \pm 0.42$ \\
 & Worst-group & $57.64 \pm 2.79$ & $53.39 \pm 2.87$ & $46.53 \pm 0.75$ & $47.37 \pm 1.95$ & $58.73 \pm 1.26$ \\
\addlinespace[2pt]
Ministral 3 & Average-group & $73.25 \pm 0.73$ & $74.24 \pm 0.84$ & $70.60 \pm 0.49$ & $72.51 \pm 0.49$ & $70.16 \pm 0.02$ \\
 & Worst-group & $55.85 \pm 0.20$ & $55.56 \pm 3.71$ & $46.65 \pm 1.05$ & $48.84 \pm 0.60$ & $58.58 \pm 1.16$ \\
\addlinespace[2pt]
Gemma 4 & Average-group & $73.59 \pm 0.59$ & $74.18 \pm 0.24$ & $71.04 \pm 0.55$ & $72.13 \pm 0.20$ & $69.89 \pm 0.79$ \\
 & Worst-group & $56.88 \pm 2.91$ & $56.30 \pm 0.81$ & $\mathbf{48.80} \pm 1.10$ & $48.40 \pm 0.83$ & $58.19 \pm 0.36$ \\
\addlinespace[2pt]
InternVL3.5 & Average-group & $74.54 \pm 0.81$ & $73.77 \pm 0.29$ & $70.43 \pm 0.53$ & $72.28 \pm 0.87$ & $70.53 \pm 0.15$ \\
 & Worst-group & $59.38 \pm 2.92$ & $55.89 \pm 0.22$ & $47.03 \pm 0.47$ & $48.46 \pm 3.06$ & $57.62 \pm 1.47$ \\
\addlinespace[2pt]
MiniCPM-V4.5 & Average-group & $74.64 \pm 0.67$ & $\mathbf{74.78} \pm 0.39$ & $70.93 \pm 0.60$ & $72.11 \pm 0.44$ & $\mathbf{71.54} \pm 0.26$ \\
 & Worst-group & $59.29 \pm 2.21$ & $\mathbf{58.16} \pm 2.93$ & $46.19 \pm 0.36$ & $47.78 \pm 1.46$ & $61.27 \pm 0.47$ \\
\midrule
\multicolumn{2}{l}{\shortstack[l]{\textbf{Rank2Replace}\\(no environment labels)}} & \multicolumn{2}{c}{Average-group: $\mathbf{80.43} \pm 0.33$} & \multicolumn{3}{c}{Worst-group: $\mathbf{65.28} \pm 0.81$} \\
\bottomrule
\end{tabular}

\end{appwidetable}

\clearpage
\subsubsection{Ranking Prompts and Replacement Schedules}
\label{app:prompt_policy}

Section~\ref{sec:ranking_replacement_design} and Fig.~\ref{fig:ranking_prompts_replacement_schedules} cross four ranking prompts with Top-preserving and two score-derived replacement schedules. All conditions share the task description, evidence criteria, slot images, and visible-evidence criterion. Queries exclude the original image and its ground-truth label. Apps.~\ref{app:ranking_templates} and~\ref{app:ranking_tasks} provide the complete prompts.

\begin{appwidetable}
\caption{Ranking costs with Qwen3-VL on an RTX 3090 (Fig.~\ref{fig:ranking_prompts_replacement_schedules}). Bold costs are column minima.}
\label{tab:ranking_costs}
\centering
\small
\setlength{\tabcolsep}{5pt}
\renewcommand{\arraystretch}{1.04}
\begin{tabular}{@{}lrrrrr@{}}
\toprule
Ranking prompt & \shortstack{Input\\tokens} & \shortstack{Output\\tokens} & Requests & \shortstack{Total time\\(hr)} & \shortstack{Mean latency\\(ms/image)} \\
\midrule
\textbf{Direct Rank} & \textbf{240,378,232} & 6,811,200 & \textbf{303,244} & \textbf{22.66} & \textbf{269} \\
Pointwise Score & 790,238,408 & \textbf{1,967,156} & 1,967,156 & 32.24 & 383 \\
Binary Selection & 766,632,536 & \textbf{1,967,156} & 1,967,156 & 31.31 & 372 \\
Sequential Selection & 1,050,504,156 & 3,327,824 & 1,663,912 & 61.18 & 726 \\
\bottomrule
\end{tabular}
\end{appwidetable}

\paragraph{Ranking cost and cache reuse.}
Direct Rank has the lowest measured ranking latency and uses no auxiliary scoring for Top-preserving replacement.

Table~\ref{tab:ranking_costs} measures ranking alone, with latency covering every query per image. For $m\geq2$ nonempty slots, Direct Rank uses one query, Pointwise Score and Binary Selection each use $m$, and Sequential Selection uses $m-1$ dependent queries. For fixed images and ranking settings, rankings are computed once and cached across epochs, seeds, hyperparameter candidates, and objectives. Previously unseen images still require ranking with the default MLP.

Score-derived schedules use training-set scores only. Matching ranking outputs are reused. Other ranking prompts require additional training-slot scoring. These scores can be shared across rows and add no validation/test scoring beyond the selected ranking prompt. Table~\ref{tab:ranking_costs} reports ranking costs rather than separate totals for every ranking--schedule combination.

\paragraph{Ranking prompts.}
Following the feature-importance query formats in LLM-Select~\cite{jeong2025llmselect}, Direct Rank orders all nonempty slot images jointly. Pointwise Score ranks slots by their expected 0--9 score, with token probabilities normalized over the ten allowed scores. Binary Selection asks whether a slot contains task evidence and ranks by the probability of Yes, normalized over Yes and No. Denote these scores for image $i$ and original slot $j$ by $s_{i,j}\in[0,9]$ and $p_{i,j}\in[0,1]$, respectively.

Sequential Selection repeatedly selects the most important remaining slot. Each query presents remaining images in original slot-ID order with the shared instructions, without earlier conversation turns or predictor feedback. The last slot is appended without a query. Empty slots are excluded from queries, assigned $s_{i,j}=p_{i,j}=0$, and placed after nonempty slots in the default fixed trailing order. Pointwise Score and Binary Selection break ties by increasing original slot ID.

\paragraph{Replacement schedules.}
For ranking prompt $a$, let $\pi_i^{(a)}$ be its slot permutation. Rank-specific training means are
\begin{equation}
\bar{s}^{(a)}_r=\frac{1}{N}\sum_{i=1}^{N}s_{i,\pi_i^{(a)}(r)},
\qquad
\bar{p}^{(a)}_r=\frac{1}{N}\sum_{i=1}^{N}p_{i,\pi_i^{(a)}(r)}.
\label{eq:rank_score_statistics}
\end{equation}
Pointwise Score uses $q_r^{(a)}=1-\bar{s}^{(a)}_r/9$; Binary Selection uses $q_r^{(a)}=1-\bar{p}^{(a)}_r$. Each heatmap row therefore retains its specified ranking even when its ranking and scoring prompts differ. Probabilities are fixed per dataset split and ranking prompt. Training means include empty-slot zeros. All conditions retain the shared donor, recipient class labels, and predictor training protocol. The default Top-preserving schedule is defined in Sec.~\ref{sec:method}.

\clearpage
\subsubsection{Donor Source}
\label{app:donor_source}

Table~\ref{tab:donor_source} in Sec.~\ref{sec:donor_source} varies donor construction with slot ranking, Top-preserving replacement, recipient class labels, and predictor training fixed. The definitions below specify the replacement embedding $u_{i,r}$ for each $r>1$.

The default shared donor uses $u_{i,r}=z_{\sigma(i),r}$ from one batch pairing per recipient. Two-donor choice samples two donors shared across ranks, then independently selects either donor's same-rank slot with equal probability. Independent donors uses a separate batch pairing at each rank. Two-donor mean sets $u_{i,r}=(z_{a_i,r}+z_{b_i,r})/2$, where both donors are shared across ranks. Two-donor mean + scale correction rescales the averaged embeddings' coordinatewise variation around the training mean before insertion.

\subsubsection{Extractor and Predictor Generalizability}
\label{app:extractor_predictor}

Section~\ref{sec:extractor_predictor} and Table~\ref{tab:extractor_predictor} report the complete comparison. We evaluate frozen FT-DINOSAUR and SPOT~\cite{Kakogeorgiou2024SPOT}, each providing seven 256-dimensional slot embeddings, with the default MLP, a linear predictor, and a Transformer. Qwen3-VL Direct Rank operates on the corresponding extractor's slot images. All three predictors use the shared training protocol and Top-preserving replacement with one shared donor.

The Transformer projects each slot embedding to 64 dimensions, prepends a learned classification token, and adds learned positional embeddings for all eight tokens. Two encoder blocks use four attention heads, a feed-forward width of 128, and zero dropout. A final layer normalization and linear output layer map the classification-token representation to $C$ logits. Positional embeddings use normal initialization with standard deviation 0.02. The classification token starts at zero, and matrix weights are independently Xavier-initialized across the encoder blocks.

\subsubsection{VLM Direct Prediction and VLM-free Inference}
\label{app:vlm_free_inference}

The macro-averaged accuracy and stage-level latency comparison appears in Sec.~\ref{sec:vlm_free_inference} and Table~\ref{tab:vlm_free_inference}.

\paragraph{Direct prediction.}
Qwen3-VL-8B-Instruct receives each full test image, the task description, and all class labels:
\begin{quote}
\small\raggedright
Task: \{task\}. Classify this image using visible evidence. Choose exactly one of these labels: \{labels\}. Return only a JSON object with key \texttt{"class"} and the chosen label; no explanation.
\end{quote}
The labels are \texttt{landbird, waterbird}, \texttt{not blond, blond}, \texttt{cat, dog}, or the 60 NICO++ class names in App.~\ref{app:ranking_tasks}. This prompt requests a label from one full image rather than a ranking of multiple slot images.

\paragraph{Permutation-invariant predictors.}
The MLP-based Deep Sets model~\cite{3294996.3295098} applies a shared $256\!\to\!64\!\to\!64$ slot MLP with ReLU after each layer, mean-pools the seven outputs, and predicts with a $64\!\to\!64\!\to\!C$ MLP with BatchNorm and ReLU after its hidden layer. Our Set Transformer variant~\cite{lee2019set} uses the Transformer above without slot positional embeddings. Both use one shared scaler per embedding coordinate, fitted over all training slots and fixed thereafter, instead of separate statistics for each slot position. Shared preprocessing and permutation-invariant aggregation remove dependence on the input slot order.

Their CE controls train on uncorrupted slots. With Rank2Replace, Qwen3-VL rankings guide training-time corruption under the default schedule, with each replacement drawn from the corresponding donor rank. Validation and test use the original slot order without replacement or VLM queries. The predictors follow the shared optimization and selection protocol, without an additional hyperparameter search. Feature-stage latency includes the shared slot MLP for Deep Sets. Set Transformer attention is included in its predictor stage. Total latency sums the stages used at inference.

\clearpage
\subsection{Additional Ablations and Analyses}
\label{app:additional_ablations}

These additional studies analyze the ranking and replacement design, then test slot count, learning rate and weight decay, ranking VLMs, presentation order, and compatibility with semantic guidance. Unless varied, they use the shared protocol in Apps.~\ref{app:evaluation} and~\ref{app:implementation}, Qwen3-VL Direct Rank, and Top-preserving replacement.

\subsubsection{Contribution of Slot Ranking}
\label{app:ranking_controls}

Table~\ref{tab:ranking_controls} varies the slot ranking under the same Top-preserving schedule. Mask-area rankings sort nonempty regions by ascending or descending area. The Qwen3-VL CE control uses ranked slots without replacement. The full combination is Rank2Replace.

\begin{appwidetable}
\caption{Contribution of slot ranking under Top-preserving replacement.}
\label{tab:ranking_controls}
\small
\setlength{\tabcolsep}{5pt}
\renewcommand{\arraystretch}{1.12}
\begin{tabular*}{\textwidth}{@{\extracolsep{\fill}}llrr@{}}
\toprule
Slot ranking & Replacement schedule & Average-group & Worst-group \\
\midrule
Random ranking & \textbf{Top-preserving} & $62.16 \pm 0.11$ & $36.14 \pm 0.84$ \\
Mask-area ascending & \textbf{Top-preserving} & $60.62 \pm 0.42$ & $32.35 \pm 1.55$ \\
Mask-area descending & \textbf{Top-preserving} & $66.75 \pm 0.35$ & $43.16 \pm 1.07$ \\
\textbf{Qwen3-VL} & None & $75.96 \pm 0.29$ & $55.48 \pm 0.63$ \\
\textbf{Qwen3-VL} & \textbf{Top-preserving} & $\mathbf{80.43} \pm 0.33$ & $\mathbf{65.28} \pm 0.81$ \\
\bottomrule
\end{tabular*}
\end{appwidetable}

With the replacement schedule fixed, Qwen3-VL rankings outperform random and mask-area orderings. Their $80.43\%$ average-group and $65.28\%$ worst-group accuracy exceed the strongest nonsemantic ordering, descending mask area, by $13.68\%$/$22.12\%$. Task-guided ordering also benefits uncorrupted prediction: Ranking + CE improves over original-order CE by $4.24\%$/$7.71\%$ (Table~\ref{tab:global_macro}), before replacement adds another $4.47\%$/$9.80\%$. Ranking thus contributes to the predictive representation as well as organizing donor replacement.

\clearpage
\subsubsection{Task-Relevant Evidence in Slot Rankings}
\label{app:ranking_evidence}

\begin{appwidefigure}
    \centering
    \includegraphics[width=\textwidth]{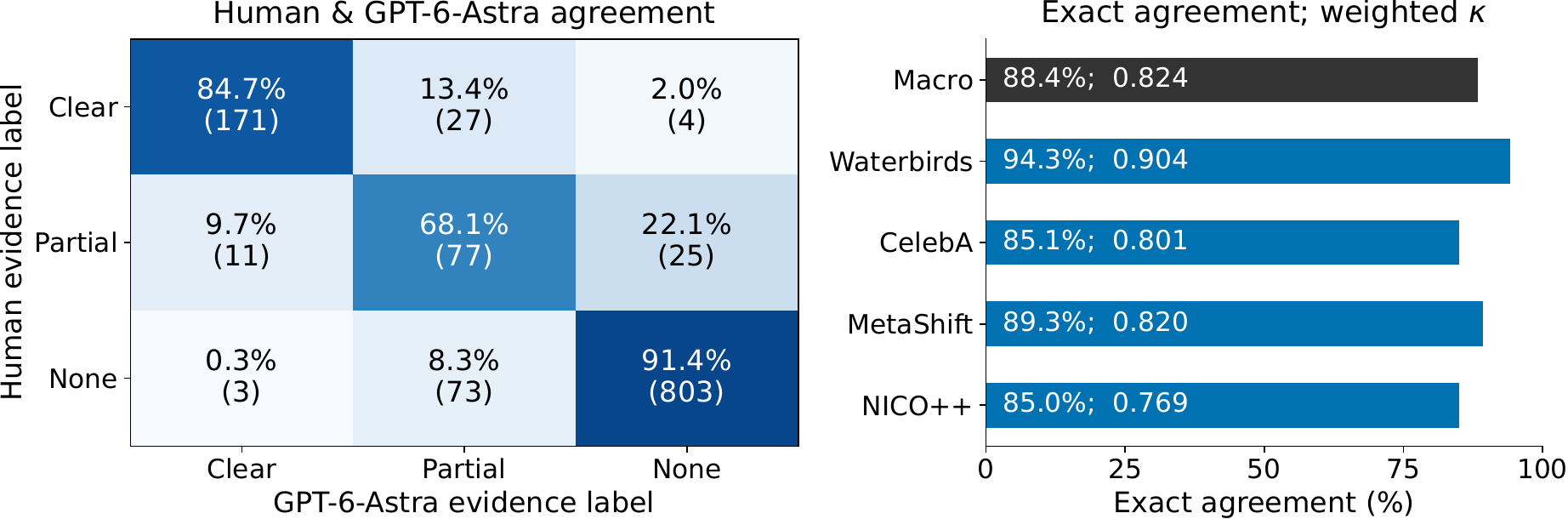}
    \par\medskip
    \includegraphics[width=\textwidth]{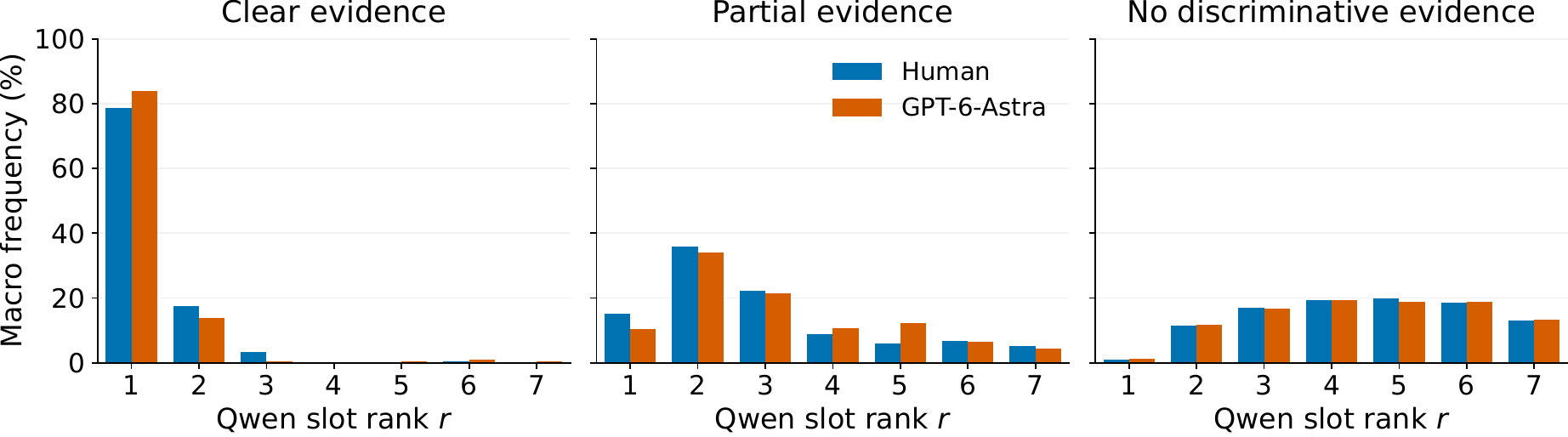}
    \caption{\textbf{Human and GPT-6-Astra evidence assessment of Qwen3-VL slot rankings.} Top: human--GPT-6-Astra agreement. Confusion-matrix cells show row-normalized percentages and pooled counts, and bars report dataset-level agreement and its macro-average. Bottom: distributions of Qwen3-VL ranks conditional on each evidence category, macro-averaged over datasets.}
    \label{fig:ranking_evidence}
\end{appwidefigure}

\paragraph{Setup.}
We assess 180 training images containing 1,194 nonempty slots. Waterbirds, CelebA, and MetaShift contribute 40 images each, with ten from every class $\times$ environment group. NICO++ contributes 60 images from the Autumn/Rock split's training partition, one per class and 15 from each of its four source environments. Qwen3-VL uses Rank2Replace's default Direct Rank prompt. A human annotator (master's degree, computer science background) and GPT-6-Astra independently assign each nonempty slot one of three evidence labels for the classification task: clear evidence, partial evidence, or no discriminative evidence. Neither annotator sees Qwen3-VL's ranks or the other's labels. We report exact agreement and linearly weighted Cohen's $\kappa$, averaged equally across the four datasets. For each annotator, we also compute rank distributions within each evidence category and dataset, then average the four distributions equally (Fig.~\ref{fig:ranking_evidence}).

\paragraph{Annotation agreement.}
Human and GPT-6-Astra labels reach $88.4\%$ macro exact agreement and a mean weighted $\kappa$ of $0.824$. Dataset-level exact agreement ranges from $85.0\%$ to $94.3\%$. Of the 143 disagreements pooled across datasets, 136 concern adjacent evidence categories. Only seven slots receive opposing clear evidence and no discriminative evidence labels.

\paragraph{Evidence placement.}
Qwen3-VL concentrates clear task evidence at the front of its rankings under both independent annotations. Across datasets, $78.74\%$ of human-labeled clear-evidence slots occupy rank one and $96.28\%$ occupy the first two ranks. GPT-6-Astra labels show the same pattern, at $83.84\%$ and $97.75\%$, respectively. Conversely, only $0.93\%$ of human-labeled slots with no discriminative evidence occupy rank one, compared with $1.27\%$ under GPT-6-Astra labels. Partial evidence is more dispersed: ranks three through seven contain $48.87\%$ under human labels and $55.45\%$ under GPT-6-Astra labels. These distributions support using task-relevance rankings to guide preservation. Partial evidence at lower ranks also motivates examining finer-grained replacement probabilities (App.~\ref{app:controlled_schedules}).

\clearpage
\paragraph{Group-level evidence retention.}
Using human labels, Table~\ref{tab:group_evidence_retention} reports image-level evidence at rank one: clear evidence (C), partial evidence (P), or no discriminative evidence (N). A ranking miss has clear evidence at another rank but not rank one. Of 26 images without clear evidence at rank one, 19 ($73.1\%$) contain no clear-evidence slot and seven ($26.9\%$) are ranking misses. When clear evidence is available, Qwen3-VL places it first in 154/161 images ($95.7\%$).

\begin{appwidetable}
\caption{Evidence retention by group. Percentages count images within each group. No C counts images with no clear-evidence slot. NICO++ images are counted once across splits. Dashes indicate no annotated images.}
\label{tab:group_evidence_retention}
\small
\setlength{\tabcolsep}{3pt}
\renewcommand{\arraystretch}{1.0}
\begin{tabular*}{\textwidth}{@{\extracolsep{\fill}}llrrrr@{}}
\toprule
Dataset & Group & $n$ & C/P/N at rank one (\%) & Ranking misses & No C \\
\midrule
Waterbirds & Landbird $\times$ Land & 10 & 80.0 / 20.0 / 0.0 & 0 & 2 \\
 & Landbird $\times$ Water & 10 & 100.0 / 0.0 / 0.0 & 0 & 0 \\
 & Waterbird $\times$ Land & 10 & 50.0 / 50.0 / 0.0 & 0 & 5 \\
 & Waterbird $\times$ Water & 10 & 90.0 / 10.0 / 0.0 & 0 & 1 \\
\midrule
CelebA & Non-blond $\times$ Female & 10 & 100.0 / 0.0 / 0.0 & 0 & 0 \\
 & Non-blond $\times$ Male & 10 & 80.0 / 10.0 / 10.0 & 1 & 1 \\
 & Blond $\times$ Female & 10 & 90.0 / 10.0 / 0.0 & 1 & 0 \\
 & Blond $\times$ Male & 10 & 100.0 / 0.0 / 0.0 & 0 & 0 \\
\midrule
MetaShift & Cat $\times$ Indoor & 10 & 100.0 / 0.0 / 0.0 & 0 & 0 \\
 & Cat $\times$ Outdoor & 10 & 90.0 / 10.0 / 0.0 & 0 & 1 \\
 & Dog $\times$ Indoor & 10 & 90.0 / 0.0 / 10.0 & 1 & 0 \\
 & Dog $\times$ Outdoor & 10 & 70.0 / 10.0 / 20.0 & 0 & 3 \\
\midrule
NICO++ & Autumn & 0 & -- & -- & -- \\
 & Dim & 15 & 73.3 / 20.0 / 6.7 & 1 & 3 \\
 & Grass & 15 & 80.0 / 13.3 / 6.7 & 2 & 1 \\
 & Outdoor & 15 & 80.0 / 0.0 / 20.0 & 1 & 2 \\
 & Rock & 0 & -- & -- & -- \\
 & Water & 15 & 100.0 / 0.0 / 0.0 & 0 & 0 \\
\bottomrule
\end{tabular*}
\end{appwidetable}

The minority groups Landbird $\times$ Water and Blond $\times$ Male retain clear evidence in all sampled images. Waterbird $\times$ Land retains clear evidence in $50\%$ and partial evidence in the rest, with no ranking misses. Thus this group's lower clear-evidence rate reflects limited clear evidence in the extracted slots. The sampled groups show no consistent minority-group disadvantage in ranking.

\paragraph{Association with prediction errors.}
Table~\ref{tab:evidence_prediction_errors} evaluates the original validation-selected checkpoints on uncorrupted ranked embeddings. The same 60 NICO++ images belong to Autumn/Rock's train partition, with 30 in test for each other split. Evidence annotation is conducted independently of model training, model selection, testing, and prompt design.

\begin{appwidetable}
\caption{Prediction errors conditional on rank-one evidence. P/N pools partial evidence and no discriminative evidence, including both ranking misses and images with no clear-evidence slot.}
\label{tab:evidence_prediction_errors}
\small
\setlength{\tabcolsep}{3pt}
\renewcommand{\arraystretch}{1.0}
\begin{tabular*}{\textwidth}{@{\extracolsep{\fill}}llrrrrrr@{}}
\toprule
Dataset split & Partition & \multicolumn{2}{c}{$n$} & \multicolumn{2}{c}{Ranking + CE error (\%)} & \multicolumn{2}{c}{Rank2Replace error (\%)} \\
\cmidrule(lr){3-4}\cmidrule(lr){5-6}\cmidrule(lr){7-8}
 & & C & P/N & C & P/N & C & P/N \\
\midrule
Waterbirds & train & 32 & 8 & $0.0\pm0.0$ & $0.0\pm0.0$ & $0.0\pm0.0$ & $0.0\pm0.0$ \\
CelebA & train & 37 & 3 & $9.0\pm3.1$ & $0.0\pm0.0$ & $17.1\pm4.1$ & $0.0\pm0.0$ \\
MetaShift & train & 35 & 5 & $0.0\pm0.0$ & $0.0\pm0.0$ & $0.0\pm0.0$ & $0.0\pm0.0$ \\
NICO++ Autumn/Rock & train & 50 & 10 & $8.0\pm2.0$ & $53.3\pm5.8$ & $8.0\pm2.0$ & $60.0\pm0.0$ \\
NICO++ Dim/Grass & train & 27 & 3 & $8.6\pm2.1$ & $66.7\pm33.3$ & $13.6\pm5.7$ & $66.7\pm0.0$ \\
 & test & 23 & 7 & $20.3\pm2.5$ & $61.9\pm21.8$ & $20.3\pm2.5$ & $76.2\pm16.5$ \\
NICO++ Outdoor/Water & train & 23 & 7 & $5.8\pm2.5$ & $23.8\pm8.2$ & $8.7\pm0.0$ & $28.6\pm14.3$ \\
 & test & 27 & 3 & $21.0\pm2.1$ & $100.0\pm0.0$ & $27.2\pm4.3$ & $88.9\pm19.2$ \\
\bottomrule
\end{tabular*}
\end{appwidetable}

P/N at rank one accompanies higher errors for both predictors on NICO++. For Dim/Grass and Outdoor/Water test images, Rank2Replace reaches $76.2\%$/$88.9\%$ error on P/N versus $20.3\%$/$27.2\%$ on C. Its P/N error changes relative to Ranking + CE are $+14.3\%$/$-11.1\%$, based on seven/three images. These results link P/N at rank one to classification difficulty, with split-dependent changes from replacement.

\clearpage
\subsubsection{Replacement Schedules}
\label{app:rulebased_schedules}
\label{app:controlled_schedules}

\paragraph{Controlled schedules and corruption content.}
Table~\ref{tab:controlled_schedules} separates the replacement probabilities from the content inserted at selected ranks. Donor schedules use whole, same-rank embeddings and uncorrupted validation/test inputs. For the non-donor controls, ranks two through seven receive zeros or synthetic embeddings while rank one is retained. Train-only masking and synthetic corruption use uncorrupted ranked inputs for validation and test, matching donor replacement. Their all-splits variants also modify these inputs.

\begin{appwidetable}
\caption{Replacement schedules and alternative corruption contents with Qwen3-VL ranking ($K=7$). Donor replacement uses uncorrupted validation/test inputs. Masking and synthetic conditions specify their corruption scope. Accuracies are macro averages (\%).}
\label{tab:controlled_schedules}
\small
\setlength{\tabcolsep}{4pt}
\renewcommand{\arraystretch}{1.12}
\begin{tabular*}{\textwidth}{@{\extracolsep{\fill}}llrr@{}}
\toprule
Setting & $(q_1,\ldots,q_7)$ & Average-group & Worst-group \\
\midrule
Ranking + CE & $(0,0,0,0,0,0,0)$ & $75.96 \pm 0.29$ & $55.48 \pm 0.63$ \\
\textbf{Top-preserving} & $\boldsymbol{(0,1,1,1,1,1,1)}$ & $\mathbf{80.43} \pm 0.33$ & $\mathbf{65.28} \pm 0.81$ \\
Top-preserving masking (train only) & $(0,1,1,1,1,1,1)$ & $60.88 \pm 0.57$ & $38.60 \pm 0.73$ \\
Top-preserving masking (all splits) & $(0,1,1,1,1,1,1)$ & $75.93 \pm 0.17$ & $55.02 \pm 0.74$ \\
Top-preserving synthetic (train only) & $(0,1,1,1,1,1,1)$ & $79.52 \pm 0.07$ & $62.15 \pm 0.38$ \\
Top-preserving synthetic (all splits) & $(0,1,1,1,1,1,1)$ & $79.38 \pm 0.12$ & $62.37 \pm 0.48$ \\
Uniform & $(1/2,1/2,1/2,1/2,1/2,1/2,1/2)$ & $71.57 \pm 0.38$ & $42.81 \pm 0.93$ \\
Linear & $(0,1/6,2/6,3/6,4/6,5/6,1)$ & $80.01 \pm 0.37$ & $63.81 \pm 1.18$ \\
Reversed lower-rank & $(0,1,5/6,4/6,3/6,2/6,1/6)$ & $79.94 \pm 0.13$ & $64.11 \pm 0.45$ \\
\bottomrule
\end{tabular*}
\end{appwidetable}

\paragraph{Moment-matched synthetic control.}
For each dataset split, we estimate the mean $\mu_{rj}$ and population standard deviation $\sigma_{rj}$ (\texttt{ddof}=0) for each coordinate $j=1,\ldots,256$ at each rank $r=2,\ldots,7$, using only ranked, standardized training embeddings. We replace these coordinates with
\begin{equation}
\widetilde z_{i,r,j}=\mu_{rj}+\sigma_{rj}\epsilon_{i,r,j},
\qquad
\epsilon_{i,r,j}\overset{\mathrm{iid}}{\sim}\mathcal{N}(0,1).
\label{eq:synthetic_slots}
\end{equation}
Sampling is independent across images, ranks, and coordinates, without preserving covariance. Coordinates with $\sigma_{rj}=0$ remain at $\mu_{rj}$. Across all six splits, standardization yields approximately zero means and unit standard deviations. No validation/test statistics or class/environment labels enter this construction.

Both synthetic conditions redraw values for every training batch. The train-only variant uses uncorrupted ranked validation/test inputs, matching the native-donor protocol. In the all-splits variant, validation and test use separate random streams for each seed, with one fixed synthetic input per image reused throughout training and final evaluation. Both variants select checkpoints by balanced accuracy plus macro F1 on their respective validation inputs. Rank-one embeddings and recipient labels remain unchanged. The MLP retains its $7\times256$ input dimensionality, and the remaining settings match the masking controls. Each synthetic condition covers six splits and three seeds, giving 18 runs.

\paragraph{Effect of corruption content.}
With uncorrupted validation/test inputs fixed, native-donor replacement exceeds train-only synthetic corruption by $0.91\%$/$3.13\%$ in average-/worst-group accuracy, and train-only masking by $19.55\%$/$26.68\%$. Train-only synthetic corruption improves on train-only zero masking by $18.64\%$/$23.55\%$, recovering much of the gap while retaining the same complete evaluation inputs. The remaining donor advantage supports native content beyond independent coordinatewise moment matching. Native donors supply jointly observed slot embeddings. Synthetic sampling removes dependencies within and across slots.

Under all-splits corruption, synthetic inputs improve on zero masking by $3.45\%$/$7.35\%$. The two synthetic protocols have similar reported means: train-only synthetic corruption gives $0.14\%$ higher average-group and $0.22\%$ lower worst-group accuracy than all-splits synthetic corruption. In comparison, all-splits masking exceeds train-only masking by $15.05\%$/$16.42\%$. These scope comparisons change both validation-based model selection and test inputs. The native-donor versus train-only synthetic comparison holds the evaluation-input protocol fixed.

\clearpage
\paragraph{Rule-based probability profiles.}
With $\ell_r=(r-1)/6$, strength scaling uses $q_r=\min(1,\alpha\ell_r)$, Above uses $q_r=1-(1-\ell_r)^\beta$, and Below uses $q_r=\ell_r^\beta$. Figure~\ref{fig:rule_based_replacement_schedules} compares $\alpha\in\{0.5,0.75,1.25,1.5\}$ and $\beta\in\{1.5,3\}$. Every profile operates on the same full task-relevance ranking, with the ranking, shared donor, and same-rank correspondence fixed.

\begin{appwidefigure}
\includegraphics[width=\textwidth]{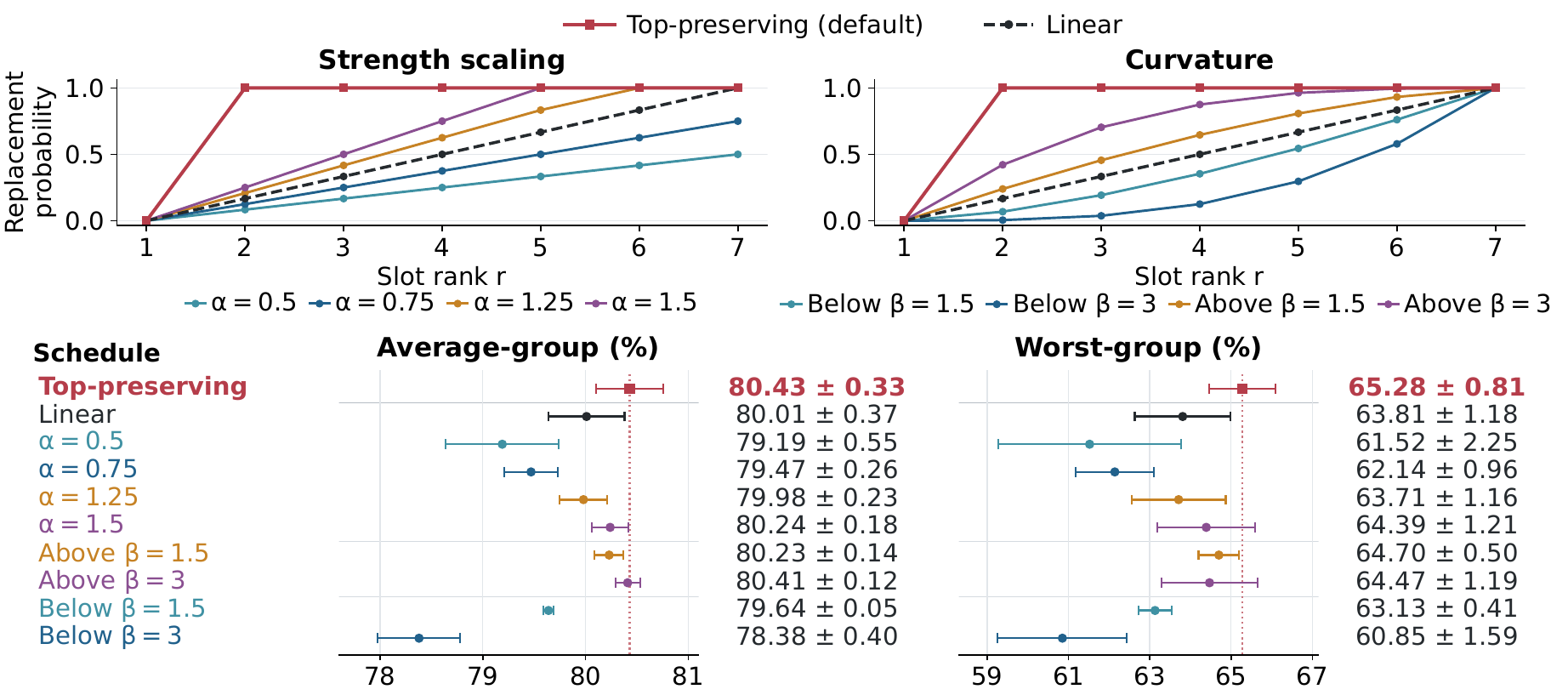}
\caption{\textbf{Rule-based replacement schedules.} Top: probability curves derived from Linear and the Top-preserving reference. Bottom: corresponding accuracies, with Qwen3-VL rankings and a shared donor fixed.}
\label{fig:rule_based_replacement_schedules}
\end{appwidefigure}

\paragraph{Probability sensitivity and the role of ranking.}
Linear and Reversed lower-rank differ by $0.07\%$ in average-group accuracy and $0.30\%$ in worst-group accuracy despite reversing probabilities below rank one. The default achieves the highest means among these rule-based profiles. These results characterize the choice of $q_r$ on fixed ranked representations: reversing a probability profile leaves both the slot permutation and donor correspondence intact.

The full task-relevance ranking organizes both prediction and replacement. With the default MLP, all recipient slots occupy rank-specific input positions at evaluation. During augmentation, rank $r$ receives the donor's embedding at the same relevance rank. A schedule determines how often that correspondence is used. VLM score-derived schedules further use rank-specific training statistics to assign graded retention probabilities, as detailed on the next page.

\clearpage
\paragraph{VLM score-derived probabilities.}
Table~\ref{tab:score_replacement_probabilities} reports probabilities derived from rank-specific training means (Eq.~\eqref{eq:rank_score_statistics}), fixed per dataset split and ranking prompt. In all 48 configurations, ranks two through four have nonzero retention probabilities. Rank-two retention ranges from $13.34\%$ to $46.65\%$. These VLM-derived profiles adjust probabilities across all ranks, including rank one.

\begin{appwidetable}
\caption{VLM score-derived replacement probabilities. Values are rounded to four decimals.}
\label{tab:score_replacement_probabilities}
\footnotesize
\setlength{\tabcolsep}{2.5pt}
\renewcommand{\arraystretch}{1.0}
\begin{tabular*}{\textwidth}{@{\extracolsep{\fill}}llrrrrrrr@{}}
\toprule
\multicolumn{9}{c}{\textbf{(a) Pointwise Score schedule}} \\
Dataset split & Ranking prompt & Rank 1 & Rank 2 & Rank 3 & Rank 4 & Rank 5 & Rank 6 & Rank 7 \\
\midrule
Waterbirds & \textbf{Direct Rank} & 0.0539 & 0.8448 & 0.9892 & 0.9978 & 0.9993 & 1.0000 & 0.9998 \\
 & Pointwise Score & 0.0489 & 0.8466 & 0.9904 & 0.9990 & 0.9998 & 1.0000 & 1.0000 \\
 & Binary Selection & 0.0519 & 0.8443 & 0.9899 & 0.9989 & 0.9998 & 1.0000 & 1.0000 \\
 & Sequential Selection & 0.0533 & 0.8448 & 0.9886 & 0.9983 & 0.9998 & 1.0000 & 1.0000 \\
\addlinespace[1.5pt]
CelebA & \textbf{Direct Rank} & 0.6307 & 0.8135 & 0.9714 & 0.9940 & 0.9981 & 0.9992 & 0.9997 \\
 & Pointwise Score & 0.6011 & 0.8313 & 0.9768 & 0.9976 & 0.9998 & 1.0000 & 1.0000 \\
 & Binary Selection & 0.6102 & 0.8266 & 0.9737 & 0.9965 & 0.9996 & 1.0000 & 1.0000 \\
 & Sequential Selection & 0.6259 & 0.8171 & 0.9704 & 0.9946 & 0.9988 & 0.9998 & 1.0000 \\
\addlinespace[1.5pt]
MetaShift & \textbf{Direct Rank} & 0.2442 & 0.8654 & 0.9855 & 0.9989 & 0.9997 & 1.0000 & 1.0000 \\
 & Pointwise Score & 0.2396 & 0.8666 & 0.9878 & 0.9997 & 1.0000 & 1.0000 & 1.0000 \\
 & Binary Selection & 0.2409 & 0.8655 & 0.9879 & 0.9994 & 1.0000 & 1.0000 & 1.0000 \\
 & Sequential Selection & 0.2477 & 0.8619 & 0.9853 & 0.9988 & 1.0000 & 1.0000 & 1.0000 \\
\addlinespace[1.5pt]
NICO++ Autumn/Rock & \textbf{Direct Rank} & 0.1753 & 0.6561 & 0.8655 & 0.9367 & 0.9638 & 0.9797 & 0.9911 \\
 & Pointwise Score & 0.1439 & 0.6104 & 0.8660 & 0.9601 & 0.9898 & 0.9982 & 0.9998 \\
 & Binary Selection & 0.1566 & 0.6147 & 0.8597 & 0.9544 & 0.9863 & 0.9971 & 0.9995 \\
 & Sequential Selection & 0.1726 & 0.6481 & 0.8618 & 0.9368 & 0.9681 & 0.9859 & 0.9949 \\
\addlinespace[1.5pt]
NICO++ Dim/Grass & \textbf{Direct Rank} & 0.1760 & 0.6710 & 0.8809 & 0.9437 & 0.9684 & 0.9816 & 0.9923 \\
 & Pointwise Score & 0.1408 & 0.6244 & 0.8852 & 0.9713 & 0.9936 & 0.9988 & 0.9998 \\
 & Binary Selection & 0.1539 & 0.6287 & 0.8778 & 0.9656 & 0.9904 & 0.9980 & 0.9996 \\
 & Sequential Selection & 0.1728 & 0.6626 & 0.8743 & 0.9456 & 0.9735 & 0.9890 & 0.9961 \\
\addlinespace[1.5pt]
NICO++ Outdoor/Water & \textbf{Direct Rank} & 0.1619 & 0.6519 & 0.8532 & 0.9232 & 0.9555 & 0.9759 & 0.9891 \\
 & Pointwise Score & 0.1284 & 0.5926 & 0.8511 & 0.9533 & 0.9877 & 0.9977 & 0.9998 \\
 & Binary Selection & 0.1420 & 0.5993 & 0.8443 & 0.9463 & 0.9831 & 0.9962 & 0.9995 \\
 & Sequential Selection & 0.1586 & 0.6452 & 0.8487 & 0.9229 & 0.9593 & 0.9821 & 0.9939 \\
\midrule
\multicolumn{9}{c}{\textbf{(b) Binary Selection schedule}} \\
Dataset split & Ranking prompt & Rank 1 & Rank 2 & Rank 3 & Rank 4 & Rank 5 & Rank 6 & Rank 7 \\
\midrule
Waterbirds & \textbf{Direct Rank} & 0.0423 & 0.8491 & 0.9899 & 0.9979 & 0.9992 & 0.9998 & 0.9999 \\
 & Pointwise Score & 0.0386 & 0.8494 & 0.9913 & 0.9989 & 0.9998 & 1.0000 & 1.0000 \\
 & Binary Selection & 0.0362 & 0.8507 & 0.9923 & 0.9990 & 0.9998 & 1.0000 & 1.0000 \\
 & Sequential Selection & 0.0413 & 0.8489 & 0.9898 & 0.9984 & 0.9996 & 1.0000 & 1.0000 \\
\addlinespace[1.5pt]
CelebA & \textbf{Direct Rank} & 0.6916 & 0.8312 & 0.9773 & 0.9964 & 0.9991 & 0.9997 & 0.9999 \\
 & Pointwise Score & 0.6717 & 0.8441 & 0.9809 & 0.9984 & 0.9999 & 1.0000 & 1.0000 \\
 & Binary Selection & 0.6661 & 0.8476 & 0.9826 & 0.9988 & 0.9999 & 1.0000 & 1.0000 \\
 & Sequential Selection & 0.6868 & 0.8348 & 0.9771 & 0.9969 & 0.9995 & 0.9999 & 1.0000 \\
\addlinespace[1.5pt]
MetaShift & \textbf{Direct Rank} & 0.2438 & 0.8607 & 0.9833 & 0.9989 & 0.9986 & 1.0000 & 1.0000 \\
 & Pointwise Score & 0.2407 & 0.8585 & 0.9866 & 0.9994 & 1.0000 & 1.0000 & 1.0000 \\
 & Binary Selection & 0.2391 & 0.8600 & 0.9866 & 0.9995 & 1.0000 & 1.0000 & 1.0000 \\
 & Sequential Selection & 0.2469 & 0.8569 & 0.9836 & 0.9979 & 1.0000 & 1.0000 & 1.0000 \\
\addlinespace[1.5pt]
NICO++ Autumn/Rock & \textbf{Direct Rank} & 0.1537 & 0.6232 & 0.8405 & 0.9168 & 0.9498 & 0.9711 & 0.9878 \\
 & Pointwise Score & 0.1276 & 0.5639 & 0.8302 & 0.9416 & 0.9836 & 0.9966 & 0.9995 \\
 & Binary Selection & 0.1158 & 0.5603 & 0.8351 & 0.9474 & 0.9869 & 0.9978 & 0.9998 \\
 & Sequential Selection & 0.1511 & 0.6155 & 0.8356 & 0.9169 & 0.9537 & 0.9782 & 0.9921 \\
\addlinespace[1.5pt]
NICO++ Dim/Grass & \textbf{Direct Rank} & 0.1602 & 0.6464 & 0.8654 & 0.9312 & 0.9589 & 0.9753 & 0.9897 \\
 & Pointwise Score & 0.1294 & 0.5878 & 0.8608 & 0.9611 & 0.9904 & 0.9981 & 0.9997 \\
 & Binary Selection & 0.1168 & 0.5849 & 0.8674 & 0.9667 & 0.9929 & 0.9988 & 0.9998 \\
 & Sequential Selection & 0.1569 & 0.6384 & 0.8567 & 0.9320 & 0.9647 & 0.9842 & 0.9943 \\
\addlinespace[1.5pt]
NICO++ Outdoor/Water & \textbf{Direct Rank} & 0.1431 & 0.6147 & 0.8232 & 0.8983 & 0.9374 & 0.9650 & 0.9849 \\
 & Pointwise Score & 0.1136 & 0.5394 & 0.8083 & 0.9304 & 0.9797 & 0.9957 & 0.9995 \\
 & Binary Selection & 0.1006 & 0.5335 & 0.8131 & 0.9382 & 0.9841 & 0.9972 & 0.9998 \\
 & Sequential Selection & 0.1395 & 0.6089 & 0.8176 & 0.8970 & 0.9409 & 0.9721 & 0.9904 \\
\bottomrule
\end{tabular*}
\end{appwidetable}

These profiles permit graded retention beyond rank one. In Fig.~\ref{fig:ranking_prompts_replacement_schedules}, score-derived schedules yield higher worst-group but lower average-group accuracy, with more expensive ranking or additional training-slot scoring (App.~\ref{app:prompt_policy}). Top-preserving remains a simple default within this accuracy--cost trade-off.

\clearpage
\subsubsection{Number of Slots}
\label{app:slot_count}

We vary FT-DINOSAUR over $K\in\{5,6,7,8,9\}$ while keeping Top-preserving replacement fixed. Rank2Replace leads both macro-averaged metrics at each tested count: average-group accuracy spans $79.83\%$--$80.90\%$, and worst-group accuracy spans $63.89\%$--$65.82\%$. Hard removal searches its retained-slot count separately for each $K$. The default replacement schedule introduces no such cutoff.

Tables~\ref{tab:slots_average} and~\ref{tab:slots_worst} provide the full numerical comparison.

\begin{appwidetable}
\caption{Average-group accuracy by slot count.}
\label{tab:slots_average}
\centering
\small
\setlength{\tabcolsep}{5pt}
\renewcommand{\arraystretch}{1.04}
\begin{tabular}{@{}lrrrrr@{}}
\toprule
Method & $K=5$ & $K=6$ & $\boldsymbol{K=7}$ & $K=8$ & $K=9$ \\
\midrule
CE & $71.86 \pm 0.24$ & $71.12 \pm 0.81$ & $71.72 \pm 0.66$ & $72.21 \pm 0.21$ & $72.29 \pm 0.25$ \\
IRM & $73.71 \pm 0.34$ & $73.12 \pm 0.34$ & $74.07 \pm 0.27$ & $73.77 \pm 0.43$ & $74.49 \pm 0.41$ \\
Group DRO & $74.95 \pm 0.08$ & $72.94 \pm 0.53$ & $73.85 \pm 0.36$ & $73.89 \pm 0.32$ & $75.15 \pm 0.22$ \\
JTT & $71.57 \pm 0.23$ & $71.25 \pm 0.29$ & $71.31 \pm 0.62$ & $71.32 \pm 0.49$ & $71.44 \pm 0.42$ \\
RSC & $71.86 \pm 0.13$ & $71.37 \pm 0.72$ & $72.23 \pm 0.65$ & $72.33 \pm 0.10$ & $72.57 \pm 0.35$ \\
CORAL & $71.46 \pm 0.42$ & $69.96 \pm 0.39$ & $70.60 \pm 0.23$ & $71.48 \pm 0.17$ & $71.22 \pm 0.02$ \\
Mixup & $71.45 \pm 0.33$ & $70.80 \pm 0.45$ & $71.79 \pm 0.77$ & $71.97 \pm 0.08$ & $72.12 \pm 0.37$ \\
ElRep & $71.98 \pm 0.24$ & $71.38 \pm 0.65$ & $71.88 \pm 0.56$ & $72.02 \pm 0.19$ & $72.43 \pm 0.43$ \\
SAGM & $72.43 \pm 0.06$ & $71.10 \pm 0.34$ & $72.06 \pm 0.19$ & $72.31 \pm 0.13$ & $72.46 \pm 0.11$ \\
SWAD & $71.41 \pm 0.07$ & $71.38 \pm 0.18$ & $71.43 \pm 0.08$ & $71.45 \pm 0.47$ & $72.21 \pm 0.15$ \\
SCER & $72.45 \pm 0.22$ & $71.30 \pm 0.41$ & $72.12 \pm 0.35$ & $72.06 \pm 0.08$ & $72.09 \pm 0.36$ \\
DFR & $69.35 \pm 0.43$ & $68.76 \pm 0.98$ & $70.12 \pm 0.42$ & $70.24 \pm 0.61$ & $70.27 \pm 0.65$ \\
\textbf{Rank2Replace} & $\mathbf{79.83} \pm 0.21$ & $\mathbf{80.90} \pm 0.12$ & $\mathbf{80.43} \pm 0.33$ & $\mathbf{80.26} \pm 0.24$ & $\mathbf{80.70} \pm 0.35$ \\
Grouped LAAT & $76.00 \pm 0.55$ & $75.71 \pm 0.53$ & $76.28 \pm 0.58$ & $75.78 \pm 0.36$ & $75.74 \pm 0.41$ \\
Hard removal & $78.89 \pm 0.32$ & $80.28 \pm 0.19$ & $79.55 \pm 0.07$ & $79.03 \pm 0.23$ & $80.00 \pm 0.33$ \\
\bottomrule
\end{tabular}

\end{appwidetable}
\begin{appwidetable}
\caption{Worst-group accuracy by slot count.}
\label{tab:slots_worst}
\centering
\small
\setlength{\tabcolsep}{5pt}
\renewcommand{\arraystretch}{1.04}
\begin{tabular}{@{}lrrrrr@{}}
\toprule
Method & $K=5$ & $K=6$ & $\boldsymbol{K=7}$ & $K=8$ & $K=9$ \\
\midrule
CE & $47.87 \pm 0.36$ & $45.43 \pm 1.76$ & $47.77 \pm 3.02$ & $49.56 \pm 1.32$ & $48.02 \pm 1.26$ \\
IRM & $57.01 \pm 1.07$ & $53.64 \pm 2.20$ & $57.64 \pm 2.79$ & $56.99 \pm 0.58$ & $57.39 \pm 1.80$ \\
Group DRO & $57.11 \pm 1.02$ & $50.26 \pm 3.92$ & $53.39 \pm 2.87$ & $55.41 \pm 0.34$ & $59.38 \pm 1.16$ \\
JTT & $44.11 \pm 1.52$ & $43.80 \pm 0.27$ & $45.62 \pm 1.69$ & $44.19 \pm 1.70$ & $44.04 \pm 2.24$ \\
RSC & $45.10 \pm 0.36$ & $43.98 \pm 1.94$ & $47.14 \pm 2.03$ & $47.93 \pm 1.50$ & $47.81 \pm 0.93$ \\
CORAL & $46.87 \pm 2.55$ & $44.79 \pm 1.98$ & $46.53 \pm 0.75$ & $49.19 \pm 0.98$ & $47.68 \pm 1.63$ \\
Mixup & $47.23 \pm 1.88$ & $44.14 \pm 1.17$ & $47.95 \pm 3.07$ & $48.85 \pm 1.24$ & $47.81 \pm 0.32$ \\
ElRep & $48.17 \pm 1.79$ & $45.80 \pm 0.37$ & $47.68 \pm 2.16$ & $49.12 \pm 1.90$ & $48.47 \pm 0.41$ \\
SAGM & $49.25 \pm 0.89$ & $44.80 \pm 1.08$ & $47.98 \pm 1.89$ & $48.99 \pm 0.87$ & $48.08 \pm 1.09$ \\
SWAD & $47.87 \pm 0.34$ & $46.60 \pm 0.90$ & $47.52 \pm 0.54$ & $47.86 \pm 2.40$ & $48.16 \pm 1.02$ \\
SCER & $48.85 \pm 1.47$ & $46.08 \pm 0.74$ & $47.37 \pm 1.95$ & $47.96 \pm 1.13$ & $47.57 \pm 0.89$ \\
DFR & $58.58 \pm 0.36$ & $56.58 \pm 1.88$ & $58.73 \pm 1.26$ & $58.77 \pm 2.70$ & $59.23 \pm 0.83$ \\
\textbf{Rank2Replace} & $\mathbf{63.89} \pm 0.80$ & $\mathbf{65.82} \pm 0.20$ & $\mathbf{65.28} \pm 0.81$ & $\mathbf{65.57} \pm 0.29$ & $\mathbf{64.93} \pm 1.06$ \\
Grouped LAAT & $54.05 \pm 1.89$ & $52.77 \pm 1.22$ & $56.21 \pm 1.49$ & $55.23 \pm 2.17$ & $55.14 \pm 1.71$ \\
Hard removal & $60.60 \pm 0.59$ & $63.76 \pm 0.21$ & $62.41 \pm 0.32$ & $60.99 \pm 1.08$ & $64.27 \pm 1.01$ \\
\bottomrule
\end{tabular}

\end{appwidetable}

\clearpage
\subsubsection{Learning Rate and Weight Decay}
\label{app:optimizer_sensitivity}

We retain the Overall Performance configurations (Sec.~\ref{sec:overall_performance}) for Rank2Replace and the three baselines with the highest macro-averaged worst-group accuracy using the original slot order: IRM, Group DRO, and DFR. We cross learning rates $\{10^{-5},10^{-4},10^{-3}\}$ with weight decays $\{0,10^{-3},10^{-1}\}$, keeping the remaining protocol fixed.

\begin{appwidefigure}
\includegraphics[width=\textwidth]{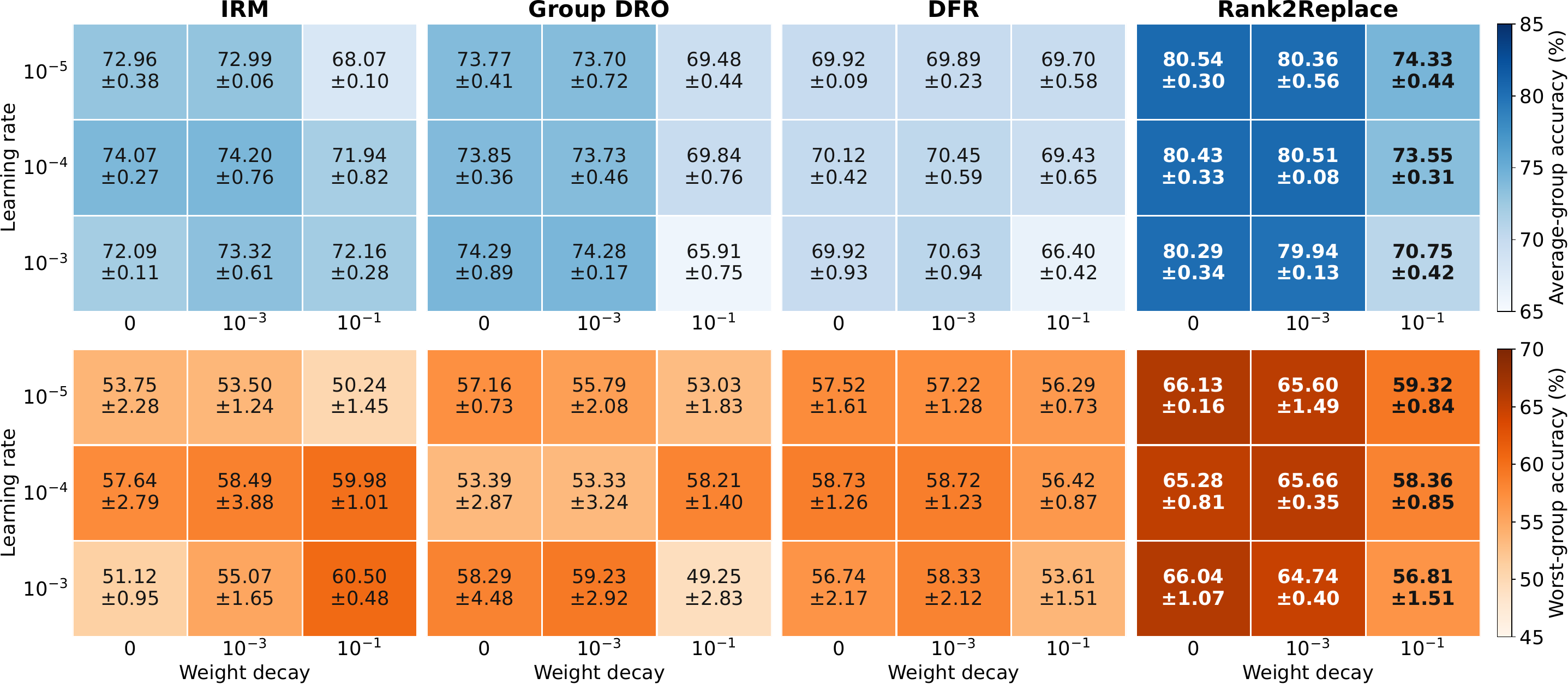}
\caption{\textbf{Learning rate and weight decay.} Each method uses the same $3\times3$ grid. Top: average-group accuracy. Bottom: worst-group accuracy. Bold annotations highlight Rank2Replace.}
\label{fig:optimizer_sensitivity}
\end{appwidefigure}

\paragraph{Performance across the sweep.}
Default Rank2Replace ($80.43\%$/$65.28\%$) exceeds every baseline configuration in Fig.~\ref{fig:optimizer_sensitivity}. Even the highest baseline average-group and worst-group means, $74.29\%$ (Group DRO) and $60.50\%$ (IRM), remain $6.14\%$ and $4.78\%$ lower. Comparing methods at matched settings, Rank2Replace leads average-group accuracy in eight of nine settings and worst-group accuracy in seven. With weight decay $0$ or $10^{-3}$, it leads both metrics at all three learning rates, spanning $79.94\%$--$80.54\%$ average-group and $64.74\%$--$66.13\%$ worst-group accuracy.

\paragraph{Learning-rate sensitivity.}
At zero weight decay, changing learning rate across two orders of magnitude changes Rank2Replace's average-group and worst-group accuracy by at most $0.25\%$ and $0.85\%$. The corresponding worst-group ranges are $6.52\%$ for IRM, $4.90\%$ for Group DRO, and $1.99\%$ for DFR. Group DRO and DFR have relatively stable average-group accuracy, with ranges of $0.52\%$ and $0.20\%$, respectively. Rank2Replace therefore maintains a broad learning-rate range with high accuracy in both metrics.

\paragraph{Weight-decay sensitivity.}
Relative to zero weight decay, $10^{-1}$ reduces both metrics for Rank2Replace and DFR at every learning rate. Its effect on IRM and Group DRO depends on learning rate. For IRM at $10^{-3}$, increasing weight decay from $0$ to $10^{-1}$ raises worst-group accuracy from $51.12\%$ to $60.50\%$ while average-group accuracy stays similar ($72.09\%$/$72.16\%$). For Group DRO, the same change raises worst-group accuracy from $53.39\%$ to $58.21\%$ at learning rate $10^{-4}$, but lowers it from $58.29\%$ to $49.25\%$ at $10^{-3}$. IRM consequently exceeds Rank2Replace in worst-group accuracy at $(10^{-4},10^{-1})$ and in both metrics at $(10^{-3},10^{-1})$. These results favor zero or light weight decay for Rank2Replace.

\clearpage
\subsubsection{Ranking VLM and Presentation Order}
\label{app:vlm_ranking}

\paragraph{Ranking VLM.}
Table~\ref{tab:ranking_vlms} compares five ranking VLMs. Table~\ref{tab:vlm_checkpoints} lists their frozen checkpoints. Each CE--Rank2Replace pair uses the same slot ranking and shared training protocol.

\begin{appwidetable}
\caption{Ranking VLMs with CE and Rank2Replace.}
\label{tab:ranking_vlms}
\centering
\small
\setlength{\tabcolsep}{7pt}
\renewcommand{\arraystretch}{1.04}
\begin{tabular}{@{}lrrrr@{}}
\toprule
 & \multicolumn{2}{c}{CE} & \multicolumn{2}{c}{\textbf{Rank2Replace}} \\
\cmidrule(lr){2-3}\cmidrule(lr){4-5}
Ranking VLM & Average-group & Worst-group & Average-group & Worst-group \\
\midrule
\textbf{Qwen3-VL} & $\mathbf{75.96} \pm 0.29$ & $\mathbf{55.48} \pm 0.63$ & $\mathbf{80.43} \pm 0.33$ & $\mathbf{65.28} \pm 0.81$ \\
Ministral 3 & $74.33 \pm 0.55$ & $52.52 \pm 1.46$ & $79.49 \pm 0.15$ & $63.25 \pm 0.48$ \\
Gemma 4 & $74.93 \pm 0.33$ & $53.45 \pm 1.36$ & $79.61 \pm 0.46$ & $63.74 \pm 1.04$ \\
InternVL3.5 & $75.33 \pm 0.23$ & $53.89 \pm 1.82$ & $79.85 \pm 0.31$ & $64.88 \pm 0.82$ \\
MiniCPM-V4.5 & $71.69 \pm 0.56$ & $46.58 \pm 2.55$ & $75.01 \pm 0.20$ & $55.99 \pm 0.64$ \\
\bottomrule
\end{tabular}

\end{appwidetable}
Slot corruption improves both metrics for every tested VLM. Gains over its ranked CE control range from $3.32\%$ to $5.16\%$ in average-group and $9.41\%$ to $10.99\%$ in worst-group accuracy. Qwen3-VL gives the highest Rank2Replace means, while absolute performance remains sensitive to the ranking VLM.

\paragraph{Slot-image presentation order.}
Table~\ref{tab:presentation_order} varies the presentation order of slot images to Qwen3-VL. Each resulting ranking orders the corresponding slot embeddings, and standardization is fitted after ranking. Default order matches the main results. Reordering 1--3 and four-order aggregation are evaluated separately.

\begin{appcompacttable}
\caption{Slot-image presentation orders for Qwen3-VL ranking.}
\label{tab:presentation_order}
\centering
\small
\setlength{\tabcolsep}{3pt}
\renewcommand{\arraystretch}{1.04}
\begin{tabularx}{\columnwidth}{@{}>{\raggedright\arraybackslash}Xrr@{}}
\toprule
Setting & Average-group & Worst-group \\
\midrule
\textbf{Default order} & $80.43 \pm 0.33$ & $65.28 \pm 0.81$ \\
Reordering 1 & $80.41 \pm 0.32$ & $65.35 \pm 0.54$ \\
Reordering 2 & $\mathbf{80.53} \pm 0.14$ & $\mathbf{65.88} \pm 0.74$ \\
Reordering 3 & $80.50 \pm 0.30$ & $65.71 \pm 1.10$ \\
Four-order aggregation & $80.47 \pm 0.09$ & $65.19 \pm 0.51$ \\
\bottomrule
\end{tabularx}

\end{appcompacttable}

Changing only presentation order produces narrower ranges: $80.41\%$--$80.53\%$ in average-group and $65.28\%$--$65.88\%$ in worst-group accuracy. Four-order aggregation reaches $80.47\%$/$65.19\%$, offering no consistent gain over the default single order.

\clearpage
\subsubsection{Compatibility with Semantic Guidance}
\label{app:semantic_compatibility}

Rank2Replace complements semantic guidance used for group construction and meta-learning. We adapt DRO-B2T~\cite{Kim_2024_CVPR} and SPUME~\cite{10.1145/3637528.3672006} to frozen FT-DINOSAUR slot embeddings. These methods introduce additional semantic supervision or training components, so we report their within-method comparisons separately from the overall comparison in Fig.~\ref{fig:baseline_performance}. Each method uses the three input conditions from Sec.~\ref{sec:ranking_compatibility}: original slot order, slot ranking, and slot ranking with slot corruption. Common optimization and model selection follow Apps.~\ref{app:implementation} and~\ref{app:evaluation}, with the method-specific settings below.

\paragraph{DRO-B2T.}
\label{app:dro_b2t}

DRO-B2T uses keyword-based CLIP text prototypes to infer a binary pseudo-bias for each training image. Crossing the class label with this pseudo-bias yields groups for Group DRO. We use the official keyword sets for Waterbirds and CelebA. For MetaShift and each NICO++ split, we fit an auxiliary CE MLP on a class-stratified 80\% of the training set and discover keywords from its predictions on the remaining 20\%. This split and keyword sampling use seed 9803. The auxiliary model's standardization uses only the 80\% subset, and its checkpoint is selected on the original validation set.

From the held-out training subset, we sample up to 128 correctly classified and 128 misclassified images per class. Qwen3-VL captions these full images without class or environment labels. YAKE extracts up to 20 keywords of at most three words from each outcome's captions. CLIP ViT-B/32 scores the combined candidates by the incorrect-minus-correct image similarity, averaged equally over classes with both outcomes. The five lowest- and five highest-scoring keywords form two sets. CLIP RN50 encodes their templated text into two normalized text prototypes, using the official 80 templates and the dataset-specific class templates for Waterbirds and CelebA. Each full training image receives the pseudo-bias of its more similar text prototype. This produces four pseudo groups for each binary task and 120 for NICO++.

Pseudo groups are fixed across input conditions and training seeds. Each condition trains a fresh MLP on the full training set using group-balanced batches and Group DRO, with group-weight step size $\eta\in\{0.001,0.01,0.1,1\}$. Slot corruption uses the default replacement schedule and shared donor, retaining the recipient's class label and pseudo group. CLIP supplies group supervision, while prediction uses slot embeddings. Ground-truth environments are used only for evaluation.

\begin{appwidetable}
\caption{DRO-B2T with cumulative slot ranking and slot corruption. Bold marks the highest mean for each metric within each row.}
\label{tab:dro_b2t_compatibility}
\small
\setlength{\tabcolsep}{3pt}
\renewcommand{\arraystretch}{1.12}
\begin{tabular*}{\textwidth}{@{\extracolsep{\fill}}lrrrrrr@{}}
\toprule
 & \multicolumn{2}{c}{Original slot order} & \multicolumn{2}{c}{+ Slot ranking} & \multicolumn{2}{c}{\textbf{+ Slot corruption}} \\
\cmidrule(lr){2-3}\cmidrule(lr){4-5}\cmidrule(lr){6-7}
Dataset / split & Average-group & Worst-group & Average-group & Worst-group & Average-group & Worst-group \\
\midrule
Macro & $74.01 \pm 0.68$ & $58.16 \pm 3.19$ & $78.06 \pm 0.35$ & $64.54 \pm 0.75$ & $\mathbf{81.75} \pm 0.10$ & $\mathbf{72.43} \pm 0.90$ \\
\midrule
Waterbirds & $76.27 \pm 0.92$ & $63.40 \pm 7.19$ & $80.73 \pm 1.15$ & $69.16 \pm 5.40$ & $\mathbf{88.40} \pm 0.21$ & $\mathbf{83.85} \pm 0.39$ \\
CelebA & $85.23 \pm 0.65$ & $62.78 \pm 3.09$ & $85.78 \pm 0.66$ & $66.11 \pm 3.85$ & $\mathbf{87.08} \pm 0.26$ & $\mathbf{72.96} \pm 1.79$ \\
MetaShift & $74.42 \pm 2.06$ & $49.31 \pm 5.93$ & $79.22 \pm 1.33$ & $59.26 \pm 3.82$ & $\mathbf{82.52} \pm 0.72$ & $\mathbf{66.67} \pm 1.39$ \\
NICO++ Autumn/Rock & $62.09 \pm 0.07$ & $61.31 \pm 0.10$ & $69.12 \pm 0.35$ & $68.06 \pm 0.31$ & $\mathbf{70.94} \pm 0.34$ & $\mathbf{70.20} \pm 0.36$ \\
NICO++ Dim/Grass & $61.12 \pm 0.61$ & $58.17 \pm 0.71$ & $67.48 \pm 0.26$ & $64.35 \pm 0.41$ & $\mathbf{69.73} \pm 0.39$ & $\mathbf{66.02} \pm 0.59$ \\
NICO++ Outdoor/Water & $57.16 \pm 0.12$ & $51.99 \pm 0.31$ & $62.87 \pm 0.21$ & $58.50 \pm 0.03$ & $\mathbf{66.32} \pm 0.22$ & $\mathbf{62.51} \pm 0.40$ \\
\bottomrule
\end{tabular*}
\end{appwidetable}

\paragraph{Results.}
Rank2Replace improves DRO-B2T in both metrics on all six splits (Table~\ref{tab:dro_b2t_compatibility}). Macro-averaged average-group and worst-group accuracies increase from $74.01\%$/$58.16\%$ with the original slot order to $78.06\%$/$64.54\%$ after ranking and $81.75\%$/$72.43\%$ after corruption. The latter step adds $3.69\%$ and $7.89\%$. On Waterbirds, worst-group accuracy rises from $63.40\%$ to $69.16\%$ to $83.85\%$. These gains show that the input intervention complements CLIP-derived group supervision.

\clearpage
\paragraph{SPUME.}
\label{app:spume}

Our SPUME adaptation uses Qwen3-VL captions of all training images, without class or environment labels. Nouns and adjectives extracted with \texttt{en\_core\_web\_sm} and occurring in more than ten training images form a fixed image--attribute matrix shared across input conditions and training seeds. Class-specific attribute spuriousness is updated each epoch from predictions on uncorrupted training inputs using SPUME's scoring rule.

Each task contains all classes. For each class, two distinct attributes are sampled according to spuriousness. Images containing the first but not the second supply support examples, and the reverse supplies query examples. Empty candidate sets trigger resampling. Each role samples ten images per class with replacement, with disjoint support and query candidates. The shared Linear--BatchNorm--ReLU block maps slot embeddings to 64-dimensional features. Support means form class prototypes, and temperature-scaled cosine similarities provide query logits. Query CE backpropagates through both query features and support prototypes. Training uses four tasks per update and 20 updates per epoch, with temperature in $\{5,10\}$ and no inner-loop optimizer.

With slot corruption, donor permutations are sampled separately within each task's support and query sets. The default replacement schedule and shared donor apply within each set, retaining recipient class labels. After each epoch, uncorrupted full-training-set features form class prototypes for validation. The selected checkpoint and its prototypes are reused for test. Validation and test inputs retain the condition's slot order without replacement. Neither task construction nor training uses environment labels.

\begin{appwidetable}
\caption{SPUME with cumulative slot ranking and slot corruption. Bold marks the highest mean for each metric within each row.}
\label{tab:spume_compatibility}
\small
\setlength{\tabcolsep}{3pt}
\renewcommand{\arraystretch}{1.12}
\begin{tabular*}{\textwidth}{@{\extracolsep{\fill}}lrrrrrr@{}}
\toprule
 & \multicolumn{2}{c}{Original slot order} & \multicolumn{2}{c}{+ Slot ranking} & \multicolumn{2}{c}{\textbf{+ Slot corruption}} \\
\cmidrule(lr){2-3}\cmidrule(lr){4-5}\cmidrule(lr){6-7}
Dataset / split & Average-group & Worst-group & Average-group & Worst-group & Average-group & Worst-group \\
\midrule
Macro & $71.75 \pm 0.79$ & $50.24 \pm 2.82$ & $74.78 \pm 0.35$ & $54.29 \pm 1.44$ & $\mathbf{80.19} \pm 0.18$ & $\mathbf{70.96} \pm 0.44$ \\
\midrule
Waterbirds & $72.83 \pm 1.55$ & $43.46 \pm 5.47$ & $76.08 \pm 0.92$ & $45.12 \pm 5.16$ & $\mathbf{85.24} \pm 0.50$ & $\mathbf{75.33} \pm 2.05$ \\
CelebA & $83.58 \pm 0.67$ & $57.59 \pm 2.51$ & $82.51 \pm 0.82$ & $53.70 \pm 3.70$ & $\mathbf{86.09} \pm 0.58$ & $\mathbf{73.89} \pm 1.67$ \\
MetaShift & $74.25 \pm 1.60$ & $46.53 \pm 4.86$ & $77.55 \pm 1.02$ & $58.10 \pm 4.07$ & $\mathbf{84.09} \pm 0.53$ & $\mathbf{71.76} \pm 0.40$ \\
NICO++ Autumn/Rock & $57.81 \pm 0.25$ & $56.97 \pm 0.26$ & $64.86 \pm 0.46$ & $64.14 \pm 0.34$ & $\mathbf{66.67} \pm 0.32$ & $\mathbf{66.42} \pm 0.22$ \\
NICO++ Dim/Grass & $57.23 \pm 0.26$ & $54.28 \pm 0.53$ & $63.36 \pm 0.27$ & $60.20 \pm 0.30$ & $\mathbf{66.09} \pm 0.24$ & $\mathbf{62.87} \pm 0.73$ \\
NICO++ Outdoor/Water & $53.96 \pm 0.30$ & $48.92 \pm 0.39$ & $60.69 \pm 0.32$ & $56.41 \pm 0.39$ & $\mathbf{63.23} \pm 0.22$ & $\mathbf{59.23} \pm 0.36$ \\
\bottomrule
\end{tabular*}
\end{appwidetable}

\paragraph{Results.}
Rank2Replace also improves SPUME in both metrics on all six splits (Table~\ref{tab:spume_compatibility}). Macro-averaged average-group and worst-group accuracies increase from $71.75\%$/$50.24\%$ to $80.19\%$/$70.96\%$. Beyond ranking alone, slot corruption adds $5.41\%$ and $16.67\%$, respectively. Its largest additional worst-group gain is on Waterbirds, from $45.12\%$ to $75.33\%$. On CelebA, corruption raises the ranked result from $53.70\%$ to $73.89\%$, exceeding the original-order result of $57.59\%$. Together with DRO-B2T, these results support compatibility with group supervision and attribute-guided meta-learning while preserving each method's training objective.

\begingroup
\fontencoding{T1}\selectfont

\clearpage
\subsection{VLM Checkpoints and Prompts}
\label{app:vlm_prompts}

\subsubsection{VLM Checkpoints}
\label{app:vlm_checkpoints}

Table~\ref{tab:vlm_checkpoints} specifies the Hugging Face checkpoints used in the slot-ranking and environment-label comparisons. Qwen3-VL is the default ranking VLM. All checkpoints remain frozen.

\begin{appwidetable}
\caption{Full Hugging Face checkpoint identifiers. Bold identifies the default VLM.}
\label{tab:vlm_checkpoints}
\centering
\small
\fontencoding{T1}\selectfont
\setlength{\tabcolsep}{5pt}
\renewcommand{\arraystretch}{1.08}
\begin{tabular}{@{}ll@{}}
\toprule
VLM & Hugging Face checkpoint \\
\midrule
\textbf{Qwen3-VL} & \textbf{\texttt{Qwen/Qwen3-VL-8B-Instruct}} \\
Ministral 3~\cite{liu2026ministral3} & \texttt{mistralai/Ministral-3-8B-Instruct-2512-BF16} \\
Gemma 4~\cite{gemmateam2026gemma4} & \texttt{google/gemma-4-E4B-it} \\
InternVL3.5~\cite{wang2025internvl3_5} & \texttt{OpenGVLab/InternVL3\_5-8B} \\
MiniCPM-V4.5~\cite{yu2025minicpmv45cookingefficient} & \texttt{openbmb/MiniCPM-V-4\_5} \\
\bottomrule
\end{tabular}
\end{appwidetable}

\subsubsection{Ranking and Scoring Templates}
\label{app:ranking_templates}

All four prompt formats prepend the common template below and then append their response instructions. The common template, task descriptions, and evidence criteria were specified from human knowledge before model training. Each query contains only nonempty slot images, without the original image or its ground-truth label. Multi-image queries prefix each image with \texttt{Region \{slot\_id\}:} and present images in increasing original slot-ID order by default. The placeholders \texttt{\{ids\}} and \texttt{\{remaining\_ids\}} list comma-separated original IDs, such as \texttt{S0,S2,S5}.

\paragraph{Common template.}
\begin{quote}
\small\raggedright
Task: \{task\}\\
Task evidence: \{evidence\}

Use only visible pixels; gray pixels are excluded. First distinguish regions containing recognizable task evidence from regions containing none. Then distinguish pure task evidence from task evidence mixed with unrelated visible content. Priority is strictly: pure task evidence, then mixed task evidence, then no task evidence. A mixed region must remain ahead of regions without task evidence, even when the unrelated content occupies more space. Several parts of the same task-relevant object remain task evidence. Within each level, prefer stronger evidence for distinguishing the possible task classes. Gray masking does not make a region mixed. Keep this assessment internal; output only the requested answer. Before treating a region as evidence, check that the task-relevant content itself is recognizable from the visible pixels. A similar color or generic texture alone is not enough. A recognizable partial view is valid evidence; a complete object is not required. Do not fill masked areas with imagined content.
\end{quote}

\paragraph{Direct Rank.}
All nonempty slot images are included in one query. Append:
\begin{quote}
\small\raggedright
These regions come from one image. Break indistinguishable ties by ascending slot ID. Return JSON with key ranking, containing exactly these IDs once, best first: \{ids\}.
\end{quote}

\paragraph{Pointwise Score.}
Each query contains one nonempty slot image. Append:
\begin{quote}
\small\raggedright
Rate this masked region from 0 to 9. Higher means stronger intrinsic evidence for identifying the target class, robust to environment changes; 0 means no useful evidence, 9 means very strong evidence. Return exactly one digit, without explanation.
\end{quote}

\paragraph{Binary Selection.}
Each query contains one nonempty slot image. Append:
\begin{quote}
\small\raggedright
Should this masked region be selected as useful intrinsic evidence for identifying the target class, robust to environment changes? Return exactly Yes or No, without explanation.
\end{quote}

\paragraph{Sequential Selection.}
Each query includes only the remaining nonempty slot images, without previous conversation turns. The last remaining slot is appended without a query. Append:
\begin{quote}
\small\raggedright
These regions come from one image. Break indistinguishable ties by ascending slot ID. Select the single most important region among these remaining IDs: \{remaining\_ids\}. Return only its ID, without explanation.
\end{quote}

\clearpage
\subsubsection{Ranking Task Descriptions and Evidence Criteria}
\label{app:ranking_tasks}

Each task description below replaces \texttt{\{task\}}, and the corresponding evidence criteria replace \texttt{\{evidence\}} in the common template. The NICO++ Autumn/Rock, Dim/Grass, and Outdoor/Water splits share the task description, evidence criteria, and all 60 class names.

\noindent\textbf{Waterbirds.}\par
\begingroup\small\raggedright
\textbf{Task:} classify bird species as waterbird versus landbird, robust to background changes\\
\textbf{Evidence:} Recognizable bird anatomy, plumage, head, beak, wings, feet and body shape. Water, vegetation, sky and other scenery alone are not bird-species evidence.
\par\endgroup\smallskip

\noindent\textbf{CelebA.}\par
\begingroup\small\raggedright
\textbf{Task:} classify whether the person has blond hair or not, robust to changes in other attributes\\
\textbf{Evidence:} Recognizable scalp hair and its visible color, including both blond and non-blond hair. Face, skin, beard, hat, clothing and background alone are not scalp-hair-color evidence.
\par\endgroup\smallskip

\noindent\textbf{MetaShift.}\par
\begingroup\small\raggedright
\textbf{Task:} classify the main animal as cat or dog, robust to indoor or outdoor environment changes\\
\textbf{Evidence:} Recognizable animal muzzle, nose, ears, eyes, fur, paws, tail and body shape that distinguish cat from dog. Furniture, rooms, grass, roads and people alone are not animal-class evidence.
\par\endgroup\smallskip

\noindent\textbf{NICO++: all three splits.}\par
\begingroup\small\raggedright
\textbf{Task:} classify the main object into one of these categories, robust to environment changes: car, flower, chair, truck, tiger, wheat, seal, wolf, lion, dolphin, lifeboat, corn, fishing rod, owl, sunflower, cow, bird, clock, shrimp, goose, airplane, rabbit, hot air balloon, lizard, hat, spider, motorcycle, tortoise, dog, crocodile, elephant, gun, fox, bus, cat, sailboat, giraffe, cactus, pumpkin, train, ship, helicopter, bicycle, racket, squirrel, bear, scooter, mailbox, horse, pineapple, frog, football, ostrich, tent, kangaroo, monkey, crab, sheep, butterfly, umbrella\\
\textbf{Evidence:} Recognizable shape, characteristic parts, material, texture and relevant color of the main object. Lighting, season, scenery, support surfaces and co-occurring objects alone are not main-object-class evidence.
\par\endgroup\smallskip

\subsubsection{Environment Annotation}
\label{app:environment_prompts}

Environment-supervised baselines use a separate prompt with the original image, resized to $224\times224$, instead of slot images. The shared template is
\begingroup\small\raggedright
Task: \{task\}. Identify the environment visible in this image. Choose one of: \{environments\}. Return JSON with key environment.
\par\endgroup\smallskip
Here, \texttt{\{task\}} is replaced by the corresponding description below, and \texttt{\{environments\}} lists the allowed categories from Table~\ref{tab:environment_prompt_categories}. For NICO++, these category sets comprise the four source environments in each split (App.~\ref{app:datasets}).

\noindent\textbf{Waterbirds.}\par
\begingroup\small\raggedright
classify bird species as waterbird versus landbird, robust to background changes; identify whether the background is land or water, not the bird type
\par\endgroup\smallskip

\noindent\textbf{CelebA.}\par
\begingroup\small\raggedright
classify blond hair versus not blond hair; the environment attribute here is the person's apparent visual gender presentation (female or male). Describe only visual presentation, not gender identity; do not use hair color as the criterion
\par\endgroup\smallskip

\noindent\textbf{MetaShift.}\par
\begingroup\small\raggedright
classify cat versus dog; identify indoor or outdoor surroundings, not the animal type
\par\endgroup\smallskip

\noindent\textbf{NICO++: all three splits.}\par
\begingroup\small\raggedright
classify the main object; identify the visual context rather than the object class. Context definitions: autumn = autumn scenery or foliage; dim = dim lighting; grass = grassy surroundings; outdoor = general outdoor surroundings; rock = rocky surroundings; water = water surroundings. Choose the best matching allowed context
\par\endgroup\smallskip

\begin{appcompacttable}
\caption{Allowed categories for environment annotation in each dataset split.}
\label{tab:environment_prompt_categories}
\centering
\small
\setlength{\tabcolsep}{3pt}
\renewcommand{\arraystretch}{1.08}
\begin{tabularx}{\columnwidth}{@{}>{\raggedright\arraybackslash}X>{\raggedright\arraybackslash}X@{}}
\toprule
Dataset split & Allowed categories \\
\midrule
Waterbirds & land, water \\
CelebA & female, male \\
MetaShift & indoor, outdoor \\
NICO++ Autumn/Rock & dim, grass, outdoor, water \\
NICO++ Dim/Grass & autumn, outdoor, rock, water \\
NICO++ Outdoor/Water & autumn, dim, grass, rock \\
\bottomrule
\end{tabularx}
\end{appcompacttable}
\endgroup

\end{document}